# Not All Synthetic Data Are Equal: Expert-Committee Audit Screening for Imbalanced Crash-Injury-Severity Prediction in Automated Driving Systems

Zewei Li[1], Qiaoqiao Ren[2], Hang Yang[1], S.C. Wong[3], Stergios-Aristoteles Mitoulis[4], Yun Ye[1,4†]

[1] Faculty of Maritime and Transportation, Ningbo University, Ningbo, China

[2] Centre for Transport Engineering and Modelling, Department of Civil and Environmental Engineering, Imperial College London, London, United Kingdom

[3] Department of Civil Engineering, The University of Hong Kong, Hong Kong, China

[4] The Bartlett School of Sustainable Construction, University College London, London, United Kingdom

† **Correspondence to**: yun.ye@ucl.ac.uk

**Acknowledgement:** This research was supported by National Natural Science Foundation of China (Project No. 72501150), Zhejiang Provincial Natural Science Foundation of China (Grant No. LQN25E080011), Ningbo Natural Science Foundation (Grant No. 2024J440), National "111" Centre on Safety and Intelligent Operation of Sea Bridges (Project No. D21013), and Research Grants Council of the Hong Kong Special Administrative Region, China (Project No. 17202824). The fourth author was also supported by Francis S Y Bong Professorship in Engineering. The funders had no role in the study design, data collection and processing, manuscript preparation, or decision to publish.

**Authorship contribution statement**

**Zewei Li:** Conceptualization, Data curation, Formal analysis, Investigation, Methodology, Software, Visualization, Writing – original draft. **Qiaoqiao Ren:** Data curation, Visualization, Writing – review & editing. **Hang Yang:** Validation, Writing – review & editing. **S.C. Wong:** Funding acquisition, Writing – review & editing. **Stergios-Aristoteles Mitoulis:** Supervision, Writing – review & editing. **Yun Ye:** Conceptualization, Methodology, Project administration, Funding acquisition, Resources, Supervision, Writing – review & editing.

**Declaration of competing interest**

The authors declare that they have no known competing financial interests or personal relationships that could have appeared to influence the work reported in this paper.

**ABSTRACT**

Automated driving systems (ADSs) are increasingly operating on public roads, raising traffic safety concerns, yet reliable prediction of crash injury severity remains challenging due to limited crash reports, rare severe outcomes, and highly imbalanced injury classes. Existing augmentation methods mainly involve increasing the minority-class sample size but rarely examine whether the generated samples are credible for safety-critical prediction. This study proposes Expert-Committee Audit Screening (ECAS), a credibility-aware sample acceptance framework for prediction of ADS-related crash injury severity under data imbalance. Using 1,477 incident-level ADS crashes from the National Highway Traffic Safety Administration Standing General Order records, ECAS audits generated minority samples through a real-data-only expert committee based on label support, boundary separation, committee agreement, and local plausibility. Within-class percentile normalization and Pareto non-dominated sorting are used to select accepted samples without manually assigned evidence weights. Under a fixed backbone combining normalizing flow augmentation and a Tabular Prior-data Fitted Network (TabPFN) classifier, the best ECAS configuration achieved the highest balanced accuracy, macro-F1, and minor-injury recall among all evidence configurations. Local neighborhood analysis confirmed that ECAS-accepted samples were better supported by nearby real minority crashes than unscreened retained samples. Shapley additive explanations and partial dependence plot analyses further indicated that the fitted model associated lower injury severity classes mainly with crash counterpart and pre-crash movement, whereas moderate-plus injuries were more sensitive to posted speed limit and operating context. These findings highlight the value of moving from quantity-oriented augmentation to credibility-aware sample acceptance for ADS safety prediction and risk governance.

## 1. Introduction

Automated driving technology is expected to improve road safety, but such benefits do not follow automatically from technological progress alone (Kalra and Paddock, 2016; Koopman and Wagner, 2017). As automated driving systems (ADSs) now operate continuously on public roads, the recording of injury severity in ADS crashes is no longer only a description of individual events. It is also becoming relevant to regulatory review, system deployment, and public trust (Zhang et al., 2024; Che et al., 2026; Ye et al., 2026a, 2026b). For traffic safety research, this further increases the importance of severity analysis. Severity does not simply represent an outcome category, but also indicates the level of risk involved, helps prioritize interventions, and guides the intensity of governance responses (Mannering et al., 2016; Schultz et al., 2025).

Against this background, research on ADS crash severity has a practical significance that differs from conventional traffic crash analysis. The National Highway Traffic Safety Administration (NHTSA) continuously collects crash reports involving ADSs and Level 2 advanced driver assistance systems (ADASs) through the Standing General Order (NHTSA, 2023). This effort is not limited to routine statistical archiving, but also supports defect identification, risk tracking, and subsequent enforcement. Compared with traditional crash databases, these records are more directly embedded in a technical regulatory context and are closer to the real-world operation of ADSs on public roads (Scanlon et al., 2024; Goodall, 2025; Ding et al., 2025). At the same time, official reporting systems adopt different definitions and reporting requirements for ADSs and Level 2 ADASs, while the relevant records continue to be revised and corrected (SAE On-Road Automated Vehicle Standards Committee, 2021). This means that empirical research on ADS crashes cannot rely on a simple transfer of general crash-severity modeling to a new setting. Rather, it requires a reexamination of injury severity formation within a risk context that has a clear institutional background and technical boundaries (Moradloo et al., 2024).

Previous studies mainly used California DMV reports or similar public records to conduct descriptive analysis and statistical modeling of basic crash characteristics, collision types, and influencing factors in automated driving crashes (Leilabadi and Schmidt, 2019; Xu et al., 2019). As the amount of available data increases, the research focus is extending to more complex questions. In addition to structured variables such as roadway environment, vehicle movement status, crash counterpart, lighting, and weather, recent studies have also examined crash narratives, latent scenario patterns, and heterogeneity across crash subgroups (Liu et al., 2024a, 2024b; Ren and Xu, 2025). This literature shows that automated driving crashes do not follow a single and stable formation path. Their outcomes are often jointly shaped by scenario conditions, system operating states, and the coupling of multiple factors (Abdel-Aty and Ding, 2024; Andriola et al., 2025). Accordingly, the research focus is gradually moving from the identification of significant variables alone toward analytical frameworks that place greater emphasis on mechanism understanding, heterogeneity, and interpretive support.

However, a more fundamental issue remains insufficiently addressed when the focus returns to severity prediction itself. For ADS crashes, real-world reported samples are limited, and severe injury events with highly consequential implications are even rarer (Huang et al., 2024; Zhu et al., 2025). Existing studies have improved the understanding of automated-driving crash risk through cost-sensitive learning, statistical modeling, text mining, explainable machine learning, and heterogeneity analysis (Ventura et al., 2024; Pan et al., 2025; Tamakloe et al., 2025). Some studies have also begun to model injury severity directly. Nevertheless, most existing methods still focus on improving classification performance, expanding minority-class samples, or identifying more important influencing factors, while giving limited attention to whether candidate samples are credible and whether they should be included in the final training set (Qian and Li, 2022). This issue is particularly important for ADS crash data, which are constrained by limited sample size, a specific reporting mechanism, and a low proportion of severe injury classes (Qiao et al., 2025). If minority-class augmentation only involves increasing the sample quantity without further checking the credibility of candidate samples, performance improvement may not correspond to more reliable identification of high-risk crashes (Sáez et al., 2015; Wang et al., 2023).

In this sense, the imbalance problem in ADS crash injury severity prediction is not only about the shortage of minority class samples (López et al., 2013; Qiu et al., 2025), but also about the credibility of candidate augmented records. While synthetic samples can help address class scarcity, they may still be located near unstable decision boundaries or in regions where real observations are extremely sparse (Barua et al., 2014; Napierala and Stefanowski, 2016). Such samples may also receive conflicting judgments from models with different inductive biases. If these samples are included in training without proper acceptance checks, augmentation may increase sample counts without improving the quality of class representation (Chen et al., 2021; Qian and Li, 2022). Although recent ADS crash studies increasingly emphasize interpretability and mechanism identification (Zhao et al., 2025), systematic checks on the reliability of generated samples are still uncommon during imbalance handling.

This study proposes Expert-Committee Audit Screening (ECAS), a synthetic sample screening framework for ADS crash injury severity prediction. The study first identifies the prediction backbone before applying sample acceptance. It then constructs an expert committee using only real samples and uses this committee as an external assessor of generated minority records. Candidate samples are examined through four audit evidence channels: label support, boundary separation, committee agreement, and local plausibility. Based on these channels, a weight-free sample acceptance mechanism combines within-class percentile normalization with Pareto non-dominated sorting (Deb et al., 2002), so that the generated minority records enter training only when they remain defensible under multiple audit views. After the final prediction framework is selected, Shapley additive explanations (SHAP) and partial dependence plot (PDP) analysis are used for post hoc interpretation of the final model. SHAP is used to examine class-specific feature contributions, while PDP places the main feature patterns on the predicted probability scale (Friedman, 2001; Lundberg and Lee, 2017; Lundberg et al., 2020). The contributions of this study are summarized as follows:

◆ **Credibility-aware augmentation:** This study shows that generated minority-class samples are not equally reliable for ADS crash injury severity prediction, even when they meet the same balancing target. It therefore shifts the process of imbalanced learning from simply adding synthetic samples to accepting credible generated records.

◆ **Expert-committee audit screening:** This study proposes ECAS, which introduces an explicit audit layer between synthetic sample generation and model training. A real-data-only expert committee evaluates candidate samples through label support, boundary separation, committee agreement, and local plausibility.

◆ **Weight-free multi-evidence sample selection:** This study develops a class-wise Pareto sample acceptance mechanism based on within-class percentile normalization and Pareto non-dominated sorting. This design avoids manually assigned evidence weights and allows multiple audit evidence channels to jointly determine which generated samples are retained for model training.

◆ **Interpretable machine learning-based analysis:** After the final prediction framework is selected, this study combines SHAP with PDP analysis to examine class-specific feature contributions and probability responses. This helps connect the predictive framework to ADS safety assessment and risk governance.

The remainder of the paper is organized as follows: Section 2 reviews related work on ADS crash severity prediction and imbalanced learning; Section 3 details the ECAS framework and evaluation methods; Section 4 describes the dataset and experimental design; Section 5 presents results and model interpretation; Section 6 discusses findings, implications, and limitations; and Section 7 concludes.

## 2. Literature Review

### 2.1 ADS crash severity prediction

Prediction research on ADS crash severity is still relatively limited, and its development has been slower than the broader literature on automated vehicle (AV) safety (Sohrabi et al., 2021). Earlier studies mainly described crash patterns and identified correlates of injury severity from manufacturer-reported records, rather than treating severity prediction as an independent modeling problem (Song et al., 2021). Wang and Li (2019) showed that severity levels were associated with liability status, roadside parking, one-way roads, roadway setting, and collision configuration. Their findings suggested that injury variation in automated driving crashes was already linked to identifiable structural conditions, rather than arising from isolated or purely incidental circumstances.

As this research stream evolved, its analytical focus became more specific. Boggs et al. (2020) extended California crash records by incorporating roadway and surrounding environmental information, and showed that both injury occurrence and rear end involvement varied systematically across operating contexts, particularly across land use settings and roadway surroundings. Zhu and Meng (2022) moved a step closer to explicit severity classification by developing a cost-sensitive classification and regression tree model, through which manufacturer, facility type, movement preceding collision, collision type, lighting condition, and crash year emerged as major determinants of AV crash severity. Although these studies did not yet establish a fully developed predictive framework, they made one point increasingly clear: severity variation in automated driving crashes is sufficiently structured to justify dedicated modeling rather than broad descriptive interpretation alone (Das et al., 2020; Fu et al., 2025).

A further change appeared when researchers began to move beyond the standard variables recorded in crash reports. Kuo et al. (2024) incorporated roadway characteristics and point of interest variables, and found that prediction performance depended not only on the classifier itself, but also on how balancing and feature selection procedures were ordered within the preprocessing pipeline. Li et al. (2024) approached the issue from a different direction and showed that crash narratives contain important severity-related information that cannot be adequately represented by structured tabular fields alone. By extracting latent topics from crash narratives and linking them to severity through explainable XGBoost, they demonstrated that themes involving vulnerable road users, lane change interactions, and crash context could materially strengthen severity discrimination. By this stage, the literature had clearly moved beyond coarse descriptive attributes and had begun to rely on richer contextual evidence.

However, the scope of studies is not entirely consistent, as some have focused specifically on ADS crashes whereas others have examined AV crashes in a broader sense or considered ADSs and ADASs jointly (Khanfar et al., 2025; Samadi et al., 2025; Wang et al., 2026). This difference in analytical scope is important when interpreting severity-related findings. Ding et al. (2024) showed that ADS crashes differ substantially from ADAS crashes in their operating environments and injury-related correlates, with ADS crashes occurring far more often in urban settings and being associated with factors such as weather, driver type, system sophistication, mileage-related indicators, and contact area.

Once this difference is acknowledged, ADS crash severity can no longer be treated as a simple extension of general AV crash analysis.

Studies have improved the understanding of ADS crash injury severity and have introduced more capable modeling and interpretation tools. Much of this literature still ends with factor identification, single-model application, or incremental performance comparison (Chen et al., 2020; He et al., 2026). Less attention has been paid to the data conditions under which ADS crash severity prediction is conducted. Reported ADS crash records remain limited, and higher injury-severity classes are sparse by nature (NHTSA, 2023). Under such conditions, the modeling challenge is not only how to select a stronger classifier, but also how to learn from sparse and highly imbalanced observations without allowing majority classes to dominate the prediction process. This issue is especially pronounced in ADS crash severity prediction, where the available records remain limited and severe injury outcomes occupy only a very small fraction of the reported crashes. It therefore becomes necessary to examine how imbalance-handling methods affect ADS crash severity prediction and, more importantly, whether the augmented minority-class samples used for model training are sufficiently reliable (Kuo et al., 2024; Channamallu et al., 2025).

### 2.2 Imbalanced learning in crash severity prediction

Small sample size and class imbalance remain persistent methodological constraints in crash severity prediction, and their effects are more pronounced in ADS-related research (Amiri et al., 2025). Higher injury-severity classes are naturally sparse, while no-injury or low-severity crashes usually dominate the observed data. Under this distribution, classifiers may achieve acceptable overall accuracy while failing to identify the most safety-relevant cases. Recent studies have further shown that imbalance is not a secondary data inconvenience, but a factor that directly shapes model behavior, feature learning, and minority-class recognition (Morris and Yang, 2021; Ali et al., 2024). Therefore, the challenge is not only to improve average predictive performance, but also to prevent the majority class from determining the learned decision structure.

Studies have addressed this problem through several lines of methods. Data-level strategies rebalance the training set through undersampling, oversampling, or interpolation-based methods such as synthetic minority oversampling (Bazarnovi and Mohammadian, 2024). Algorithm-level strategies modify the learning objective, for example by assigning larger costs or weights to rare but consequential severity classes (Wan and Zhu, 2022; Wan et al., 2025). More recent studies have also introduced transfer learning or generative augmentation to enrich minority-class representation under data-limited conditions (Li et al., 2023; Chen et al., 2024). These methods are useful because they explicitly recognize that conventional accuracy-based training can obscure weak recognition of rare severity classes. However, they do not fully resolve the reliability problem associated with augmented samples.

A key limitation of conventional oversampling is that it improves class balance mainly in terms of sample counts. Random oversampling can repeatedly expose the model to the same minority observations, which may increase overfitting and amplify noise already present in the original data (Batista et al., 2004; Meng and Li, 2022). Interpolation-based methods reduce exact duplication, but they still rely on local neighborhood assumptions (Chawla et al., 2002; Fernández et al., 2018). When minority samples are sparse or located close to majority-class regions, interpolated observations may fall near unstable decision boundaries or in areas where the true class structure is uncertain (López et al., 2013; Sáez et al., 2015; García et al., 2020). In this situation, augmentation can make the training set appear more balanced while introducing samples that weaken boundary clarity (Soltanzadeh and Hashemzadeh, 2021).

Generative augmentation provides a more flexible alternative, but it raises a deeper question of credibility control (Zhao et al., 2021; Li et al., 2026). Generative models can produce new observations that resemble the training distribution at an aggregate level, yet this does not guarantee that each generated sample is locally plausible, correctly supported by real observations, or consistently recognized by different classifiers (D'souza et al., 2025; Hernandez et al., 2025). This issue is particularly important in crash severity analysis because rare severe outcomes are not only statistically scarce, but also safety-critical. A synthetic sample may carry a minority-class label, but still be located in a low-density region, close to a competing class, or in a part of the feature space where model predictions are unstable (Prati et al., 2004; García et al., 2020). If such samples are directly used for model training, augmentation may increase minority-class *volume* without improving minority-class *representation*.

This limitation suggests that imbalanced learning in ADS crash injury severity prediction should move beyond the question of how many minority samples can be generated. A more important question is whether candidate augmented samples are reliable enough to support model training. Existing studies have often treated generated or resampled observations as valid once they are produced, and rarely included explicit sample-level checks before incorporating them into the final training set. This gap motivates ECAS, an expert-committee audit screening framework that evaluates candidate synthetic samples through label support, boundary separation, committee agreement, and local plausibility before they are accepted for final model training.

### 2.3 Summary

Despite the growing use of data augmentation techniques to address class imbalance, a fundamental issue remains largely overlooked. Most existing approaches implicitly assume that generated or resampled observations are valid representations of minority classes. However, in safety-critical applications such as automated-driving crash analysis, this assumption may not hold. Synthetic samples can introduce noise, distort class boundaries, or fail to capture the true underlying distribution, especially when the original data are extremely limited.

More importantly, few studies have incorporated explicit mechanisms to evaluate the reliability or credibility of augmented samples before using them for model training. As a result, performance improvements achieved through augmentation may be accompanied by reduced robustness or misleading inference. This limitation suggests that, beyond rebalancing class distributions, there is a need for a systematic framework that can assess and filter augmented data based on their quality and consistency with observed data.

## 3. Methodology

### 3.1 Expert-Committee Audit Screening framework

ECAS changes the unit of decision in imbalanced ADS crash learning. Standard augmentation asks how many minority samples should be added, whereas the proposed framework first asks whether a generated record is credible enough to enter model training. This distinction matters because higher-consequence crashes are sparse, and even a small number of weak synthetic records can reshape the learned decision boundary. Generated minority samples are therefore held in a candidate pool, assessed by a real-data-only expert committee, and accepted through class-wise Pareto sample selection.

The workflow consists of three phases. First, raw ADS crash records are processed through feature engineering, data splitting, and candidate backbone comparison, with the prediction backbone selected using inner validation balanced accuracy and the one standard error rule. Second, a redundant synthetic candidate pool is generated under the fixed backbone and audited by a real data only expert committee. Candidate samples are assessed using four evidence channels: label support, boundary separation, committee agreement, and local plausibility. Accepted samples are then retained through class-wise percentile normalization and Pareto ranking. Third, the selected ECAS evidence configurations are evaluated on a holdout test set, followed by local neighborhood quality checks and class-specific model interpretation using SHAP and partial dependence profiles. The framework shifts imbalanced crash severity learning from quantity-oriented augmentation to credibility-aware sample acceptance[1]. Fig. 1 summarizes the framework.

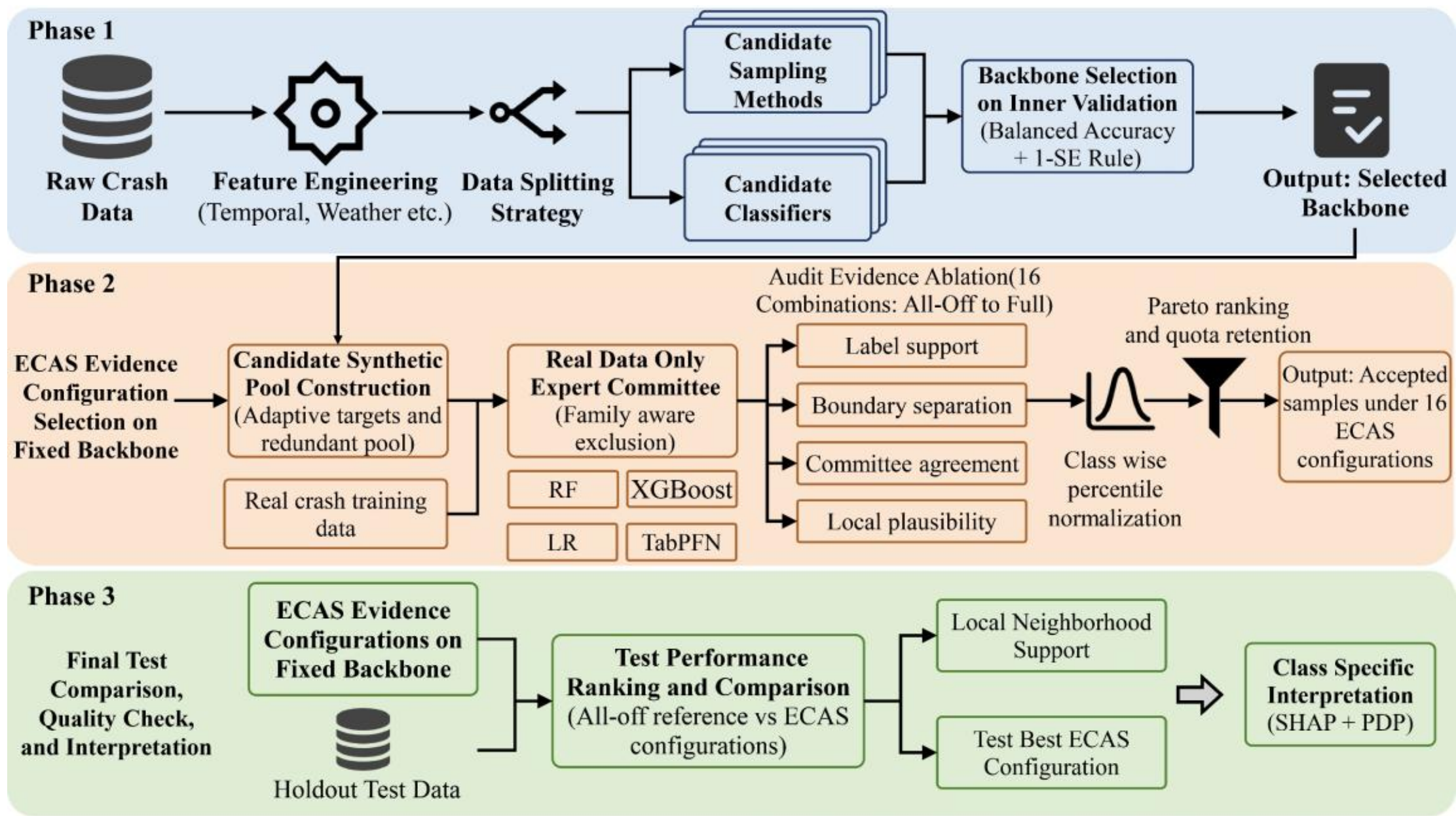

**Fig. 1.** Expert-Committee Audit Screening framework for credibility-aware synthetic sample acceptance in ADS crash injury severity prediction.

### *3.1.1 Backbone identification*

[1] Source code available on github: https://github.com/lizewei6666-star/expert-committee-audit-screening/tree/main

Let the full sample set be denoted as $D$, and let the severity class set be denoted as $C = \{1, 2, \dots, c\}$. For each seed, the data are first partitioned into an outer training subset $D_{tr}^{(s)}$ and an outer test subset $D_{te}^{(s)}$. The training subset is then further divided into a subtraining subset $D_{sub}^{(s)}$ and an inner validation subset $D_{val}^{(s)}$. Backbone identification and evidence configuration selection are both completed exclusively within $D_{sub}^{(s)}$ and $D_{val}^{(s)}$, whereas $D_{te}^{(s)}$ is used only once for the final independent evaluation. To reduce the influence of accidental data splits, in this study, the whole procedure is performed for five replicates with five random seeds, namely $\{42, 52, 62, 72, 82\}$, and the mean and standard deviation are reported.

In this study, a backbone is defined as the combination of an oversampling generator and a downstream classifier. The candidate oversampling set is denoted as

$$G = \{\text{SMOTE}, \text{ADASYN}, \text{NF}, \text{CTGAN}, \text{TVAE}\} \tag{1}$$

and the candidate classifier set is denoted as

$$M = \{\text{RF}, \text{ET}, \text{HGB}, \text{LR}, \text{XGB}, \text{CatBoost}, \text{LGBM}, \text{TabPFN}\} \tag{2}$$

The generator set includes the synthetic minority over-sampling technique (SMOTE; Chawla et al., 2002), adaptive synthetic sampling (ADASYN; He et al., 2008), normalizing flow (NF; Rezende and Mohamed, 2015), conditional tabular generative adversarial network (CTGAN; Chen et al., 2024), and tabular variational autoencoder (TVAE; Xu et al., 2019). The classifier set includes random forest (RF; Breiman, 2001), extra trees (ET; Geurts et al., 2006), histogram gradient boosting (HGB; Friedman, 2001), logistic regression (LR; Cox, 1958), extreme gradient boosting (XGB; Chen and Guestrin, 2016), categorical boosting (CatBoost; Prokhorenkova et al., 2018), light gradient boosting machine (LightGBM; Ke et al., 2017), and tabular prior-data fitted network (TabPFN; Hollmann et al., 2025).

Accordingly, a backbone can be written as

$$B = (g, m), \quad g \in G, m \in M \tag{3}$$

At this stage, all synthetic samples are incorporated into the training set according to the adaptive class target, without sample screening. This setting identifies the most competitive backbone under an uncontrolled augmentation condition and provides a stable reference for the subsequent evidence ablation stage.

A common practice in traditional imbalanced learning is to manually specify target sample sizes for different minority classes (Adeel et al. 2024, Mostafa et al. 2025). In crash injury severity prediction, however, these quotas are often empirical balancing choices for a particular dataset rather than targets grounded in the underlying severity structure (Sun et al. 2007). To avoid this problem, this study does not use manually defined class-specific targets. Instead, the augmentation target is generated automatically from the real class distribution in the current training fold.

Let $n_c$ denote the number of real samples in class $c$ within the training set. Let $M$ denote the majority class, and let $n_M$ denote its sample size. The final target size $t_c$ is defined as

$$t_c = \begin{cases} n_M, & c = M \\ \min\left\{n_M, \max\left(n_c, \left\lceil \sqrt{n_c n_M} \right\rceil\right)\right\}, & c \neq M \end{cases} \tag{4}$$

Equation (4) uses the geometric mean, $\sqrt{n_c n_M}$, as the expansion anchor. This design allows extremely rare classes to receive meaningful augmentation, while avoiding the direct expansion of all classes to the majority class size. By construction, $n_c \leq \left\lceil \sqrt{n_c n_M} \right\rceil \leq n_M$: the target size is neither smaller than the original class size nor larger than the majority class size. It therefore provides a balance between conservativeness and compensation.

Generating only the exact number of samples required by the final target is not sufficient, because screening requires room for sample acceptance. A larger candidate pool is therefore constructed. The candidate pool target for each class is defined as

$$p_c = \begin{cases} n_M, & c = M \\ \min\left\{n_M, \max\left(t_c, \left\lceil \sqrt{t_c n_M} \right\rceil\right)\right\}, & c \neq M \end{cases} \tag{5}$$

It is clear that $p_c \geq t_c$. The ECAS layer therefore operates on a redundant candidate set rather than on a set whose size is equal to the final quota. This ensures that the subsequent sample acceptance step has practical meaning. The number of synthetic samples to be added for class $c$ is further denoted as $q_c = \max(0, t_c - n_c)$.

For each backbone, an augmented training set is first constructed on $D_{sub}^{(s)}$ according to Eq. (4). The classifier $m$ is then fitted, and performance metrics are calculated on $D_{val}^{(s)}$. Balanced accuracy ($BalAcc$) is used as the primary selection metric and is defined as

$$\mathrm{BalAcc} = \frac{1}{C}\sum_{c=1}^{C}\mathrm{Recall}_c = \frac{1}{C}\sum_{c=1}^{C}\frac{TP_c}{TP_c + FN_c} \tag{6}$$

Here, $TP_c$ and $FN_c$ denote the true positives and false negatives of class $c$, respectively. This metric explicitly reduces the dominance of majority classes over the overall evaluation and is more suitable for multi-class imbalanced settings.

Suppose that the $BalAcc$ values of a backbone over the five random seeds are $\mathrm{BalAcc}_B^{(1)},\ldots,\mathrm{BalAcc}_B^{(S)}$, where $S = 5$. The mean and standard error are then calculated as

$$\overline{\mathrm{BalAcc}}_B = \frac{1}{S}\sum_{s=1}^{S}\mathrm{BalAcc}_B^{(s)} \tag{7}$$

$$\mathrm{SE}_B = \frac{\mathrm{SD}\left(\mathrm{BalAcc}_B^{(1)},\ldots,\mathrm{BalAcc}_B^{(S)}\right)}{\sqrt{S}} \tag{8}$$

Let $B^*$ denote the backbone with the highest mean $BalAcc$. According to the one standard error rule, the candidate set is defined as

$$R_B = \left\{B : \overline{BalAcc}_B \geq \overline{BalAcc}_{B^*} - SE_{B*}\right\} \tag{9}$$

Within $R_B$, this study further compares mean accuracy and macro-F1. When the performances of two backbones are close, the more stable backbone is preferred.

*3.1.2 Candidate pool construction and expert committee*

After the backbone is selected, the generated samples enter the candidate pool for audit screening. The screening layer needs to remain relatively independent from the final prediction backbone. It should also avoid allowing a model from the same family to directly validate the candidate samples used by the final predictor. For this reason, this study first defines a baseline expert pool composed of representative models with different inductive mechanisms:

$$T_0 = \{\mathrm{LR}, \mathrm{RF}, \mathrm{XGB}, \mathrm{TabPFN}\} \tag{10}$$

In this pool, LR represents a linear discrimination perspective, RF represents a bagging-based tree ensemble perspective, XGB represents a boosting-based tree ensemble perspective, and TabPFN represents a pretrained tabular foundation model perspective. This design emphasizes the heterogeneity of evidence sources rather than the number of expert models.

To further reduce family overlap between the backbone and the expert committee, this study introduces a family mapping function $\phi(\cdot)$. The model families are defined as

$$F_{\text{linear}} = \{\text{LR}\}, \quad F_{\text{bag}} = \{\text{RF}, \text{ET}\} \tag{11}$$

$$F_{\text{boost}} = \{\text{HGB}, \text{XGB}, \text{LightGBM}, \text{CatBoost}\}, \quad F_{\text{prior}} = \{\text{TabPFN}\} \tag{12}$$

If the classifier in the selected backbone is denoted as $m^*$, its corresponding family representative is defined as

$$\phi(m^*) = \begin{cases} \text{LR}, & m^* \in F_{\text{linear}} \\ \text{RF}, & m^* \in F_{\text{bag}} \\ \text{XGB}, & m^* \in F_{\text{boost}} \\ \text{TabPFN}, & m^* \in F_{\text{prior}} \end{cases} \tag{13}$$

Accordingly, the active expert committee is defined as

$$T(m^*) = T_0 \setminus \{\phi(m^*)\} \tag{14}$$

This means that when the backbone belongs to a given model family, the representative expert from that family is removed from the expert pool. This step reduces dependence between the audit screening layer and the final predictor.

The expert committee is trained only on real training samples and does not use any synthetic samples. Let $X_{real}$ and $y_{real}$ denote the features and labels of real samples in $D_{sub}^{(s)}$, respectively. For each expert model in $T(m^*)$, five-fold stratified training is performed on the real data. If $h_{kr}$ denotes the *r*-th expert model trained in the *k*-th fold, the expert committee can be written as

$$H = \{h_{kr} \mid k = 1, \ldots, K; r = 1, \ldots, R^*\} \tag{15}$$

where $K = 5$ and $R^* = |T(m^*)|$. In ECAS, "expert" refers to an external assessor trained on real crash records, not a model that directly supervises the final classifier.

*3.1.3 Audit evidence channels*

The expert committee produces four audit evidence channels for each candidate sample. These channels correspond operationally to confidence, margin, consistency, and density, but they are used here as evidence for sample acceptance rather than as isolated evaluation indicators. Let the candidate synthetic sample be denoted as $x_i$, and let its generated label be $y_i^{syn}$. For expert $h$ in the committee $H$, the posterior probability assigned to class $c$ is denoted as $p_i^{(h)}(c)$.

(1) Label support (confidence)

The committee-averaged posterior probability is defined as

$$\bar{p}_i(c) = \frac{1}{H}\sum_{h=1}^{H} p_i^{(h)}(c) \tag{16}$$

The label support evidence, operationalized by confidence, is then given by

$$a_i^{\text{conf}} = \bar{p}_i(y_i^{syn}) \tag{17}$$

This evidence measures the average support given by the expert committee to the generated label. A larger $a_i^{\text{conf}}$ indicates that the candidate sample is closer to its target class in probabilistic terms.

(2) Boundary separation (margin)

A high target-class probability does not necessarily imply a clear class boundary. To capture boundary separation, the margin evidence is defined as

$$a_i^{\text{mar}} = \bar{p}_i\left(y_i^{\text{syn}}\right) - \max_{c \neq y_i^{\text{syn}}} \bar{p}_i(c) \tag{18}$$

This quantity measures the advantage of the target class over the strongest competing class. A larger $a_i^{\text{mar}}$ suggests that the candidate sample is located in a more clearly separated decision region.

(3) Committee agreement (consistency)

Let $\hat{y}_i^{(h)}$ denote the hard prediction made by expert $h$ for sample $x_i$. The committee agreement evidence, operationalized by consistency, is defined as

$$a_i^{\text{con}} = \frac{1}{H}\sum_{h=1}^{H} \mathbb{I}\left(\hat{y}_i^{(h)} = y_i^{\text{syn}}\right) \tag{19}$$

where $\mathbb{I}(\cdot)$ is the indicator function. This evidence reflects the discrete agreement of the expert committee on the generated label. Unlike confidence, which relies on averaged probabilities, consistency focuses on whether the experts reach the same class decision.

(4) Local plausibility (density)

For crash severity prediction, a candidate sample may receive strong classification support but still deviate from the local structure of real samples. To control this issue, this study introduces local plausibility evidence, operationalized by density. Let $\bar{d}_i$ denote the average Euclidean distance from candidate sample $x_i$ to its $k$ nearest neighbors in the reference set, with $k = 15$. Let $Q_{0.05}$ and $Q_{0.95}$ denote the 5th and 95th percentiles of the average distance distribution over all candidate samples. The density evidence is defined as

$$a_i^{\text{den}} = 1 - \text{clip}\left( \frac{\bar{d}_i - Q_{0.05}}{Q_{0.95} - Q_{0.05}}, 0, 1 \right) \tag{20}$$

Here, $clip(\cdot)$ truncates values outside the interval $[0,1]$. A larger $a_i^{den}$ indicates that the candidate sample is closer to a locally dense region of the real sample distribution.

### *3.1.4 Multi-objective ranking and ablation experiment*

All four audit evidence channels are defined so that larger values indicate stronger candidate credibility. The channels differ in scale, distribution range, and variability, so direct aggregation through fixed linear weights would introduce another empirical choice. The method therefore applies percentile normalization within each class for every enabled evidence channel. For the *j*-th evidence channel in class $c$, the within-class percentile rank of candidate sample $x_i$ is defined as

$$r_{ij}^{(c)} = \frac{\text{rank}\left(a_{ij}^{(c)}\right) - 1}{N_c - 1}, \qquad r_{ij}^{(c)} \in [0,1] \tag{21}$$

Here, $N_c$ denotes the number of candidate synthetic samples in class $c$. This transformation changes the comparison from an absolute comparison across classes to a relative comparison within the same class. As a result, extremely rare classes are not placed at a systematic disadvantage simply because their audit evidence values are lower in absolute scale.

Let the currently enabled evidence channel set be $A \subseteq \{cond, mar, con, den\}$. For two candidate samples $x_u$ and $x_v$ from the same class, if

$$r_{uj}^{(c)} \geq r_{vj}^{(c)}, \qquad \forall j \in A \tag{22}$$

and there exists at least one dimension $j^* \in A$ such that

$$r_{uj^*}^{(c)} > r_{vj^*}^{(c)} \tag{23}$$

then $x_u$ is said to dominate $x_v$ in the Pareto sense. Accordingly, the candidate samples of class $c$ can be decomposed into the first front, the second front, and subsequent successive fronts. The first front consists of all candidates that are not dominated by any other sample. After this front is removed, the same procedure is repeated for the remaining samples until every candidate is assigned to a Pareto level.

When several samples belong to the same Pareto front, their average percentile rank is used for within-front ordering:

$$\bar{r}_i^{(c)} = \frac{1}{|A|}\sum_{j\in A} r_{ij}^{(c)} \tag{24}$$

Within the same front, a larger $\bar{r}_i^{(c)}$ indicates higher priority. In this way, all candidate samples in class $c$ are arranged into a complete ordered sequence. The acceptance rule first follows the Pareto front level and then uses the average percentile rank to break ties within the same front.

For class $c$, only the top $q_c$ candidate samples are retained. The final dataset used for model training is therefore written as

$$D_{fit}^{(s)} = D_{real}^{(s)} \cup D_{syn,keep}^{(s)} \tag{25}$$

To construct a strictly comparable uncontrolled reference, this study also defines an all-off configuration. In this setting, all four audit evidence channels are disabled.

The switch combination of the four audit evidence channels is defined as

$$S = \left\{ (z_{conf}, z_{mar}, z_{con}, z_{den}) \middle| z_j \in \{0,1\} \right\} \tag{26}$$

This produces 16 audit evidence configurations. For each configuration, training and validation are conducted under the same fixed backbone, active expert committee, and candidate pool. Because the candidate pool is shared by all 16 configurations under the same seed, differences among configurations arise only from the sample acceptance rule rather than from random variation in the generation process.

The 16 configurations separate the effect of each evidence channel from the effect of the generator and classifier. At this stage, the comparison defines how candidates are accepted under different evidence structures. External test performance is evaluated only after the backbone, candidate pool, and acceptance rule are fixed for each seed.

#### *3.1.5 Final external evaluation*

After the backbone is fixed, all 16 audit evidence configurations are further evaluated on the outer test set. This step examines the practical generalization ability of different evidence structures on unseen data. For each random seed, the candidate synthetic sample pool is first reconstructed from the outer training set, and the expert committee is trained using only real training samples. Each evidence configuration then follows the same procedure of sample acceptance, model training, and external testing. In this way, the performance of different configurations can be compared under identical test conditions.

### 3.2 Explainable artificial intelligence

After the final prediction framework is determined, explainable artificial intelligence is used to examine how the model forms its injury severity predictions. In this study, SHAP and PDP analysis are used as complementary interpretation tools (Friedman, 2001; Lundberg and Lee, 2017). SHAP explains how observed feature values contribute to class-specific predictions, whereas PDP analysis describes how the model's average predicted probability changes when a selected feature is systematically varied.

SHAP is derived from the Shapley value in cooperative game theory and provides an additive decomposition of model output (Shapley, 1953). For a trained model, let $f_k(x_i)$ denote the predicted output of sample $x_i$ for injury class $k$. The SHAP representation is written as

$$f_k(x_i) = E\left[f_k(x)\right] + \sum_{j=1}^{p} \phi_{ij}^{(k)} \tag{27}$$

Here, $E\left[f_k(x)\right]$ denotes the baseline prediction for class $k$, and $\phi_{ij}^{(k)}$ denotes the contribution of feature $j$ to the prediction of sample $i$ for class $k$. A positive $\phi_{ij}^{(k)}$ increases the predicted tendency toward class $k$, whereas a negative value reduces it.

To measure global feature importance, this study calculates the mean absolute SHAP value of each feature for each injury class:

$$I_j^{(k)} = \frac{1}{n}\sum_{i=1}^{n} \left|\phi_{ij}^{(k)}\right| \tag{28}$$

A larger $I_j^{(k)}$ indicates that feature $j$ has a stronger average influence on the prediction of class $k$. Based on this ranking, the most influential variables for each injury class are further examined through SHAP main-effect analysis. This analysis shows how observed feature values affect model output and helps identify which conditions increase or decrease the predicted tendency toward each injury severity class.

PDP analysis is introduced as a complementary interpretation method (Greenwell, 2017). Unlike SHAP, which explains contributions for observed samples, PDP analysis examines the average response of the trained model when one feature is set to different values. For feature $j$ and injury class $k$, the partial dependence function is defined as

$$PD_j^{(k)}\left(v\right) = \frac{1}{n}\sum_{i=1}^{n} f_k\left(v, x_{i,-j}\right) \tag{29}$$

In this expression, $v$ denotes a selected value of feature $j$, and $x_{i,-j}$ denotes all other features of sample $i$. During this calculation, feature $j$ is fixed at $v$ for all samples, while the remaining features retain their observed values. The resulting profile shows how the average predicted probability of class $k$ changes with feature $j$.

Most variables in this study are categorical or grouped variables. PDP analysis is therefore implemented in a categorical form. For a selected feature $j$, each observed category $v$ is substituted into the test samples in turn. The modified samples are then passed through the same preprocessing and prediction pipeline, and the average predicted probability of each injury class is calculated. This treatment avoids interpreting category codes as continuous numerical scales and keeps the PDP analysis consistent with the data structure of ADS crash records.

SHAP and PDP analysis answer related but different questions. SHAP identifies the contribution of observed feature values to class-specific predictions, whereas PDP analysis shows the model's average probability response under controlled changes in a selected feature. Used together, they provide a fuller interpretation of the final ADS crash severity model.

### 3.3 Evaluation indicator

To evaluate the proposed framework, this study uses a set of metrics that describe model performance from complementary perspectives. The first perspective concerns overall classification performance. The second concerns balanced recognition across injury severity classes. The third focuses on minority-class recognition, which is central to crash severity prediction under class imbalance. Because ADS-crash injury-severity levels are highly imbalanced across classes, overall accuracy alone may provide an incomplete assessment. A model can achieve a high accuracy by predicting the dominant class well, while still failing to identify rare but safety-critical severity classes. Therefore, accuracy, balanced accuracy, macro averaged F1 score (macro-F1), and minority recall are used as the main evaluation metrics. Their mean values and standard deviations are reported over repeated experiments with multiple random seeds to reduce the influence of a single random split.

For the $K$-class injury severity prediction task, the prediction results are summarized by a confusion matrix $M = \left[M_{ij}\right] \in R^{K \times K}$. The off-diagonals $M_{ij}$ record the number of observations that belong to class $i$ but are predicted as class $j$. That is, the elements $M_{ij}$, with $i \neq j$, represent classification errors between injury severity classes. Correct predictions are located on the main diagonal, where $M_{ii}$ denotes the number of correctly identified samples in class $i$. The evaluation metrics are defined based on this matrix.

(1) Balanced accuracy ($BalAcc$)

$BalAcc$ is used as the primary comparison metric because it measures the average recognition ability across classes. It is defined as

$$\text{BalAcc} = \frac{1}{K}\sum_{c=1}^{K}\text{Recall}_c \tag{30}$$

The recall of class $c$ is calculated as

$$\text{Re}\,call_c = \frac{M_{cc}}{\sum_{j=1}^{K} C_{cj}} \tag{31}$$

$BalAcc$ assigns the same weight to the recall of each class. This property is important for imbalanced crash severity data because it prevents the evaluation from being dominated by the most frequent class. In this study, $BalAcc$ therefore serves as the main basis for comparing competing models and evidence configurations.

(2) Accuracy

Accuracy measures the proportion of correctly classified samples among all test observations. It is defined as

$$Accuracy = \frac{\sum_{c=1}^{K} M_{cc}}{N} \tag{32}$$

where $N$ denotes the total number of test samples. This metric provides a direct measure of overall classification performance. However, when the data are strongly imbalanced, accuracy may mainly reflect the model's ability to predict the majority class. It is therefore reported as a supplementary metric rather than as the sole criterion for model comparison.

(3) Macro-F1

Macro-F1 is used to jointly evaluate precision and recall under multi-class imbalance. The precision of class $c$ is defined as

$$Precision_c = \frac{M_{cc}}{\sum_{i=1}^{K} M_{ic}} \tag{33}$$

The F1 score of class $c$ is calculated as

$$F1_c = \frac{2 \cdot Precision_c \cdot Recall_c}{Precision_c + Recall_c} \tag{34}$$

Macro-F1 is then obtained by averaging the class-specific F1 scores:

$$Macro - F1 = \frac{1}{K}\sum_{c=1}^{K} F1_c \tag{35}$$

Unlike accuracy, macro-F1 gives equal importance to all classes and considers both false positives and false negatives. It is therefore useful for assessing whether the model maintains balanced classification quality across different injury severity classes.

(4) Minority recall

Minority recall is introduced to directly evaluate the model's ability to identify rare injury severity classes. Let $C_{\min}$ denote the set of minority classes. Minority recall is defined as

$$\text{Minority Recall} = \frac{1}{|C_{\min}|}\sum_{c \in C_{\min}} \text{Recall}_c \tag{36}$$

This metric focuses on the average recall of minority classes and is closely aligned with the purpose of the proposed framework. A higher minority recall indicates that the model is more capable of identifying low-frequency and safety-critical injury outcomes, rather than only improving performance for the dominant class.

The reported metrics separate average correctness from class balance and minority-class recognition. Accuracy describes general prediction correctness, $BalAcc$ gives equal weight to each injury severity class, macro-F1 combines precision and recall across classes, and minority recall focuses on the sparse injury classes. This design is necessary because high overall accuracy can still mask poor recognition of injury classes that carry greater safety relevance.

# 4. Dataset

## 4.1 Data source

The crash records were drawn from the ADS subset of the archived NHTSA Standing General Order 2021-01 reporting data (NHTSA, 2023). The source file used in this study was SGO-2021-01_Incident_Reports_ADS.csv, obtained from the NHTSA Standing General Order crash reporting portal. It contains ADS incident reports with crash months from July 2021 through June 2025. Under the order, specified manufacturers and operators must report qualifying crashes involving vehicles equipped with an ADS or Level 2 ADAS. The present study uses only the ADS reports, keeping the analytical sample within a single automation domain.

The archived SGO ADS file records reports rather than unique crashes, so the unit of analysis was rebuilt before model estimation. For repeated submissions under the same Report ID, the record with the highest Report Version was retained. The remaining records were consolidated by *Same Incident ID*, which served as the primary incident identifier. Candidate duplicate groups were checked against a small set of harmonized crash descriptors, including incident timing, crash counterpart, pre-crash movements of the crash partner and subject vehicle, and injury severity. This process reduced the raw file from 2,295 report rows to 1,520 incident records. After records with invalid injury labels or missing variables required for estimation were removed, the final analytical sample contained 1,477 ADS crashes.

After report-level cleaning, incident-level deduplication, and variable harmonization, the analytical sample contains 1,477 ADS crashes. Fourteen variables are used as predictors. The dependent variable is constructed from the NHTSA SGO field Highest Injury Severity Alleged, which reports the highest confirmed or alleged injury severity for each incident. The original injury severity levels are grouped into three classes: *No injury*, *Minor injury*, and *Moderate-plus injury*. *Moderate*, *Serious*, and *Fatality* are combined into the *Moderate-plus injury* class because records in the higher severity levels are sparse. The explanatory variables are grouped into four domains: temporal characteristics, roadway and environmental characteristics, crash interaction characteristics, and background characteristics.

**4.2 Data description**

Table 1 presents the descriptive statistics of the final ADS crash sample. The final sample includes 1,477 ADS crashes. The three injury severity classes are markedly imbalanced, with no-injury crashes accounting for the clear majority of observations and the other two injury categories comprising much smaller shares. The sample is concentrated in urban traffic environments. Streets and intersections are the predominant roadway settings. Most crashes occurred under dry surface conditions, normal road conditions, daylight or dark-lighting conditions, and clear weather, and light vehicles are the most common crash counterpart. In terms of pre-crash movement, the crash partner most often traveled straight, whereas the ADS-equipped subject vehicle was most often stopped or parked; straight travel is the second-most common pre-crash movement for the subject vehicle. The background variables further indicate that most retained crashes involved relatively new vehicles, moderate to high mileage levels, and posted speed limits of 25–35 mph. These descriptive results show that the ADS crashes retained for analysis occurred mainly in routine urban operating environments with low to moderate posted speed limits.

**Table 1**. Descriptive statistics of the ADS crash dataset

| Variable | Description | Count | Percentage (%) |
|---|---|---|---|
| Month | Jan | 117 | 7.92 |
| | Feb | 126 | 8.53 |
| | Mar | 140 | 9.48 |
| | Apr | 154 | 10.43 |
| | May | 183 | 12.39 |
| | Jun | 87 | 5.89 |
| | Jul | 89 | 6.03 |
| | Aug | 127 | 8.60 |
| | Sep | 113 | 7.65 |
| | Oct | 118 | 7.99 |
| | Nov | 101 | 6.84 |
| | Dec | 122 | 8.26 |
| Day type | Weekday | 1,002 | 67.84 |
| | Weekend | 475 | 32.16 |
| Hour | Early morning | 207 | 14.01 |
| | Morning | 369 | 24.98 |
| | Afternoon | 516 | 34.94 |
| | Evening | 385 | 26.07 |
| Roadway type | Street | 849 | 57.48 |
| | Intersection | 464 | 31.42 |

| Variable | Description | Count | Percentage (%) |
|---|---|---|---|
| | Parking lot | 98 | 6.64 |
| | Highway/freeway | 59 | 3.99 |
| | Other | 7 | 0.47 |
| Road surface | Dry | 1,393 | 94.31 |
| | Wet | 81 | 5.48 |
| | Other | 3 | 0.20 |
| Road condition | Normal | 1,406 | 95.19 |
| | Work zone | 22 | 1.49 |
| | Traffic incident | 10 | 0.68 |
| | Other | 39 | 2.64 |
| Lighting condition | Daylight | 914 | 61.88 |
| | Dark-lighted | 499 | 33.78 |
| | Dawn/Dusk | 43 | 2.91 |
| | Dark-unlighted | 18 | 1.22 |
| | Other | 3 | 0.20 |
| Weather condition | Clear | 1,191 | 80.64 |
| | Cloudy | 228 | 15.44 |
| | Rain | 54 | 3.66 |
| | Other | 4 | 0.27 |
| Crash counterpart | Light vehicle | 897 | 60.73 |
| | Heavy vehicle | 297 | 20.11 |
| | Vulnerable user | 74 | 5.01 |
| | Fixed object | 46 | 3.11 |
| | Other | 163 | 11.04 |
| Crash-partner movement | Straight | 594 | 40.22 |
| | Backing/parking | 196 | 13.27 |
| | Lane change/passing | 186 | 12.59 |
| | Turning | 124 | 8.40 |

| Variable | Description | Count | Percentage (%) |
|---|---|---|---|
| | Departure/opposing | 75 | 5.08 |
| | Stopped/parked | 58 | 3.93 |
| | Other | 244 | 16.52 |
| Subject-vehicle movement | Straight | 457 | 30.94 |
| | Backing/parking | 22 | 1.49 |
| | Lane change/passing | 35 | 2.37 |
| | Turning | 114 | 7.72 |
| | Departure/opposing | 11 | 0.74 |
| | Stopped/parked | 802 | 54.30 |
| | Other | 36 | 2.44 |
| Mileage | 0–1,000 miles | 58 | 3.93 |
| | 1,000–10,000 miles | 306 | 20.72 |
| | 10,000–50,000 miles | 662 | 44.82 |
| | >50,000 miles | 451 | 30.53 |
| Vehicle age | 0–2 years | 1,086 | 73.53 |
| | 3–5 years | 310 | 20.99 |
| | ≥6 years | 73 | 4.94 |
| | Unknown | 8 | 0.54 |
| Posted speed limit | 0–20 mph | 224 | 15.17 |
| | 25–35 mph | 1,066 | 72.17 |
| | 40–45 mph | 129 | 8.73 |
| | ≥50 mph | 58 | 3.93 |
| Severity | No injury | 1,336 | 90.45 |
| | Minor injury | 109 | 7.38 |
| | Moderate-plus injury | 32 | 2.17 |

**4.3 Experimental design**

To ensure leakage-free and comparable evaluation, each experiment follows the same nested partitioning procedure. The full sample is first divided into a training set and an outer test set at a 70%/30% ratio. The outer test set remains untouched until the final evaluation and is not used for backbone identification or evidence configuration comparison. The training set is then further divided into a subtraining set and an inner validation set at a 70%/30% ratio. As a result, the subtraining set, inner validation set, and outer test set account for 49%, 21%, and 30% of the full sample, respectively. All preprocessing operations, oversampling procedures, expert-committee training, and ECAS-based sample acceptance are fitted only within the corresponding training data of each split. Details of the encoding scheme, model-specific scaling, distance-based evidence computation, and SHAP/PDP preprocessing workflow are provided in Appendix A. The experiments are repeated over five random seeds, and the results are summarized using the mean and standard deviation of each evaluation metric.

All experiments are implemented in a consistent Python environment, with data preprocessing, model estimation, ECAS-based sample acceptance, and performance evaluation conducted through widely used machine learning and deep learning libraries. The computational platform consists of an Intel Core i7-13700KF CPU and an NVIDIA GeForce RTX 4070 Ti GPU. This information is reported to make the experimental setting transparent and reproducible. Because all candidate configurations follow the same data partitioning scheme, preprocessing protocol, and evaluation metrics, the comparison focuses on predictive performance and ECAS sample acceptance rather than hardware-dependent computational differences.

## 5. Results

The results follow the logic of the framework. Inner holdout performance first identifies the common backbone. The fixed backbone then isolates ECAS sample acceptance from model choice, with external test results evaluating prediction and local neighborhood support examining the accepted synthetic records themselves. SHAP and PDP profiles finally connect the selected model to class-specific injury severity patterns.

### 5.1 Backbone selection

The inner validation results favor a balanced rather than single metric choice of backbone. ADASYN with LGBM gives the highest accuracy, at 88.45%, while CTGAN with HGB gives the highest *BalAcc*, at 37.71% (Table 2). NF augmentation with a TabPFN classifier remains close to the best class-balanced result, with an accuracy of 87.68% and a *BalAcc* of 37.15%. Under the one standard error rule with *BalAcc* as the primary criterion, NF with TabPFN stays within the eligible set and offers the most stable tradeoff between overall correctness and class-balanced recognition. The subsequent ECAS ablation and external test comparison therefore use NF-TabPFN as the fixed backbone.

**Table 2.** Validation performance of the best backbone under each oversampling method

| Oversampling method | Downstream classifier | Acc | BalAcc |
|---|---|---|---|
| NF | TabPFN | 87.68% | 37.15% |
| CTGAN | HGB | 87.03% | 37.71% |
| SMOTE | LR | 84.65% | 36.96% |
| ADASYN | LGBM | 88.45% | 36.10% |
| TVAE | LR | 83.03% | 35.03% |

**5.2 Effects of ECAS sample acceptance**

Fixing NF-TabPFN separates the effect of sample acceptance from the effect of model choice. The external test comparison evaluates the 16 evidence configurations under the same outer splits. The neighborhood analysis then examines whether the retained synthetic minority samples are locally supported by real crashes from the same injury class.

*5.2.1 Test performance across evidence configurations*

Under the fixed NF-TabPFN backbone, the best external test result comes from *audit_ c1m1s0d1*, which enables confidence, margin, and density while disabling consistency (Table 3). *BalAcc* reaches 0.3765 $\pm$ 0.0230 and macro-F1 reaches 0.3777 $\pm$ 0.0288, compared with 0.3578 $\pm$ 0.0216 and 0.3575 $\pm$ 0.0270 under the all-off reference. Minor-injury recall also rises from 0.1212 $\pm$ 0.0525 to 0.1758 $\pm$ 0.0583. Since the backbone, candidate pool construction, and outer test splits are held constant, these differences isolate the contribution of ECAS sample acceptance.

The ordering of configurations clarifies the roles of the audit evidence channels. Margin appears in all of the top five configurations and in seven of the top eight, indicating that boundary separation is the most influential evidence channel in this experiment. This finding is consistent with the screening design: a credible synthetic sample should not only receive support for its assigned class, but should also be distinguishable from its closest competing class. Confidence and density contribute to the best-ranked configuration, suggesting that label support and local plausibility provide complementary information. Consistency does not improve the leading configuration. The full evidence setting, *audit_ c1m1s1d1_full*, ranks second, but its accuracy, *BalAcc*, macro-F1, and minor-injury recall are all lower than those of *audit_ c1m1s0d1*.

The results also show that performance is not determined by the number of enabled evidence channels alone. Adding more evidence can make acceptance stricter, but it may also remove useful candidates or increase the influence of a less informative channel. The comparison between *audit_ c1m1s0d1* and *audit_ c1m1s1d1_full* illustrates this point. When consistency is added to the confidence margin density configuration, accuracy decreases from 0.8743 to 0.8716, *BalAcc* from 0.3765 to 0.3736, macro-F1 from 0.3777 to 0.3745, and minor-injury recall from 0.1758 to 0.1697. A different pattern is observed for *audit_ c0m0s1d0*, which records the highest accuracy in the table, 0.8761 ± 0.0116, but ranks only 14th because its *BalAcc* and minor-injury recall are lower. This contrast reinforces the need to interpret model performance through class-balanced and class-specific measures rather than through overall correctness alone.

**Table 3**. Test performance of ECAS evidence configurations

| **Rank** | **Confidence** | **Margin** | **Consistency** | **Density** | **Accuracy** | **Balanced accuracy** | **Macro-F1** | **Minor injury recall** |
|---|---|---|---|---|---|---|---|---|
| 1 | 1 | 1 | 0 | 1 | 0.8743 ± 0.0151 | 0.3765 ± 0.0230 | 0.3777 ± 0.0288 | 0.1758 ± 0.0583 |
| 2 | 1 | 1 | 1 | 1 | 0.8716 ± 0.0175 | 0.3736 ± 0.0255 | 0.3745 ± 0.0316 | 0.1697 ± 0.0628 |
| 3 | 0 | 1 | 1 | 0 | 0.8734 ± 0.0129 | 0.3706 ± 0.0249 | 0.3706 ± 0.0294 | 0.1576 ± 0.0691 |

| Rank | Confidence | Margin | Consistency | Density | Accuracy | Balanced accuracy | Macro-F1 | Minor injury recall |
|---|---|---|---|---|---|---|---|---|
| 4 | 1 | 1 | 0 | 0 | 0.8730 ± 0.0146 | 0.3704 ± 0.0254 | 0.3706 ± 0.0307 | 0.1576 ± 0.0691 |
| 5 | 1 | 1 | 1 | 0 | 0.8725 ± 0.0132 | 0.3702 ± 0.0245 | 0.3700 ± 0.0284 | 0.1576 ± 0.0691 |
| 6 | 1 | 0 | 1 | 0 | 0.8725 ± 0.0154 | 0.3702 ± 0.0255 | 0.3704 ± 0.0309 | 0.1576 ± 0.0691 |
| 7 | 0 | 1 | 0 | 0 | 0.8721 ± 0.0113 | 0.3701 ± 0.0242 | 0.3694 ± 0.0275 | 0.1576 ± 0.0691 |
| 8 | 0 | 1 | 0 | 1 | 0.8716 ± 0.0107 | 0.3699 ± 0.0134 | 0.3699 ± 0.0169 | 0.1576 ± 0.0332 |
| 9 | 1 | 0 | 0 | 0 | 0.8716 ± 0.0155 | 0.3699 ± 0.0258 | 0.3700 ± 0.0312 | 0.1576 ± 0.0691 |
| 10 | 0 | 1 | 1 | 1 | 0.8685 ± 0.0138 | 0.3687 ± 0.0197 | 0.3682 ± 0.0238 | 0.1576 ± 0.0498 |
| 11 | 1 | 0 | 0 | 1 | 0.8680 ± 0.0132 | 0.3686 ± 0.0145 | 0.3679 ± 0.0183 | 0.1576 ± 0.0332 |
| 12 | 1 | 0 | 1 | 1 | 0.8676 ± 0.0144 | 0.3684 ± 0.0145 | 0.3676 ± 0.0185 | 0.1576 ± 0.0332 |
| 13 | 0 | 0 | 1 | 1 | 0.8689 ± 0.0129 | 0.3670 ± 0.0186 | 0.3665 ± 0.0217 | 0.1515 ± 0.0479 |

| Rank | Confidence | Margin | Consistency | Density | Accuracy | Balanced accuracy | Macro-F1 | Minor injury recall |
|---|---|---|---|---|---|---|---|---|
| 14 | 0 | 0 | 1 | 0 | 0.8761 ± 0.0116 | 0.3660 ± 0.0290 | 0.3665 ± 0.0338 | 0.1394 ± 0.0819 |
| 15 | 0 | 0 | 0 | 1 | 0.8608 ± 0.0131 | 0.3585 ± 0.0150 | 0.3564 ± 0.0164 | 0.1333 ± 0.0346 |
| 16 (All-off) | 0 | 0 | 0 | 0 | 0.8689 ± 0.0156 | 0.3578 ± 0.0216 | 0.3575 ± 0.0270 | 0.1212 ± 0.0525 |

Figure 2 gives the test ranking a visible structure. Most configurations lie in a narrow *BalAcc* band of about 0.368 to 0.371. The selected setting, *audit_ c1m1s0d1*, sits above this band, while the all-off reference anchors the lower end. The separation identifies where screening matters: weakly supported generated records are less likely to define the augmented training set, and candidates backed by the enabled evidence channels are retained instead. With the backbone and test splits fixed, the figure points to sample acceptance as the source of the performance difference. The fact that *audit_ c1m1s0d1* outperforms the full evidence setting suggests that the most useful audit structure is not necessarily the strictest one. Confidence, margin, and density represent three complementary acceptance requirements: the generated sample should receive probabilistic support for its assigned class, should be sufficiently separated from the nearest competing class, and should remain close to a locally supported region of the real crash distribution. By contrast, consistency is based on hard expert decisions and therefore may discard probability information already captured by confidence and margin. In sparse minority-class regions, requiring discrete agreement among all experts can also become overly conservative, because plausible boundary samples may be rejected simply because different models express uncertainty in different ways. This explains why adding consistency to the confidence-margin-density configuration slightly reduces Balanced Accuracy, Macro-F1, and Minor injury recall. Therefore, the best configuration should be interpreted not as evidence that consistency is useless, but as evidence that hard agreement may be less informative than probabilistic support, boundary separation, and local plausibility in this ADS crash dataset.

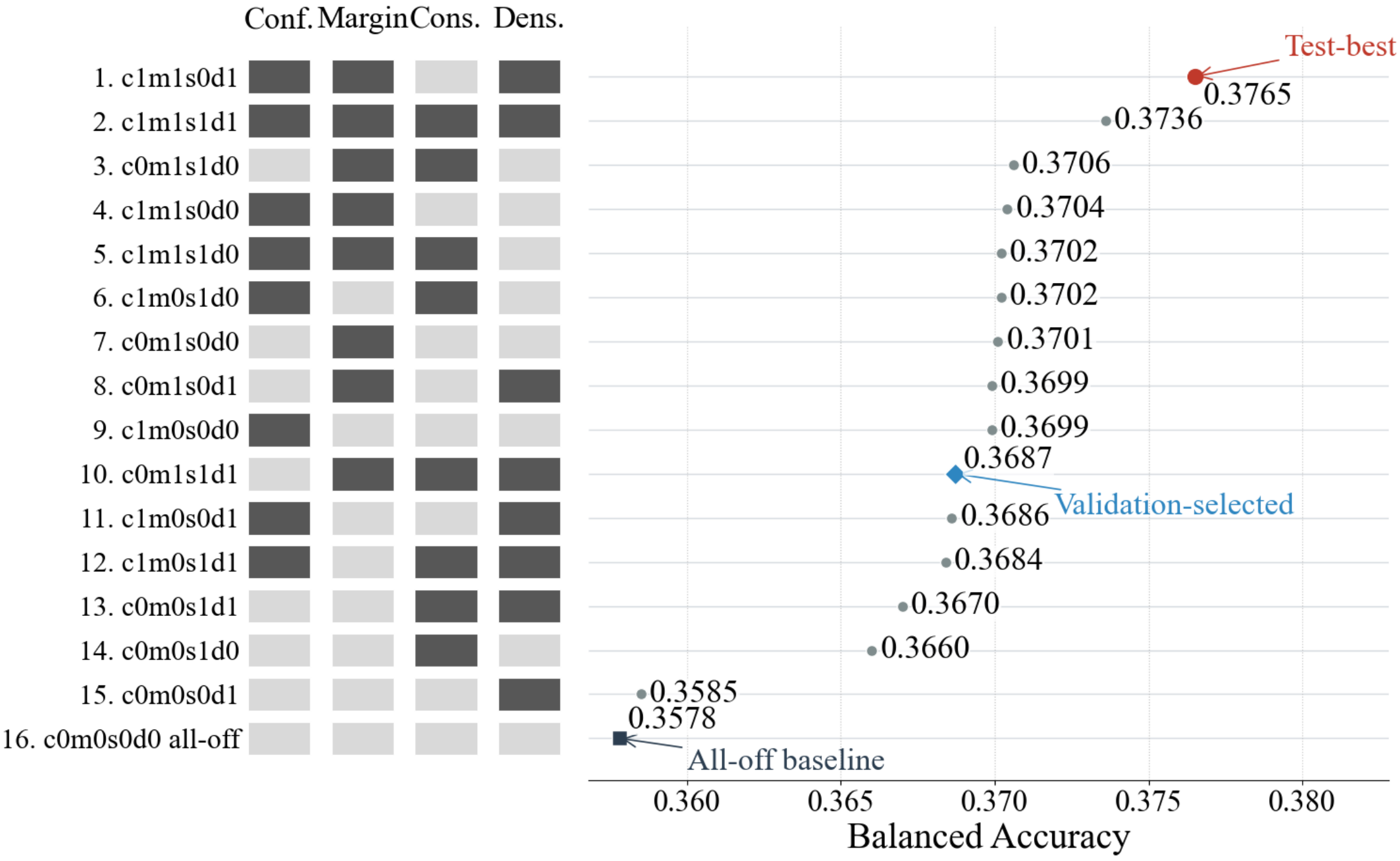

**Fig. 2.** Balanced accuracy ranking of the 16 ECAS evidence configurations. Note: c, m, s, and d denote confidence, margin, consistency, and density evidence, respectively; 1 = enabled and 0 = disabled.

*5.2.2 Local neighborhood support*

The value of ECAS becomes clearer when the accepted samples are examined directly. Figure 3(a) reports the same-class neighbor ratio, and Fig. 3(b) reports the same-class $k$ nearest neighbor (kNN) distance. A higher ratio and a lower distance indicate stronger support from real crashes of the same injury class.

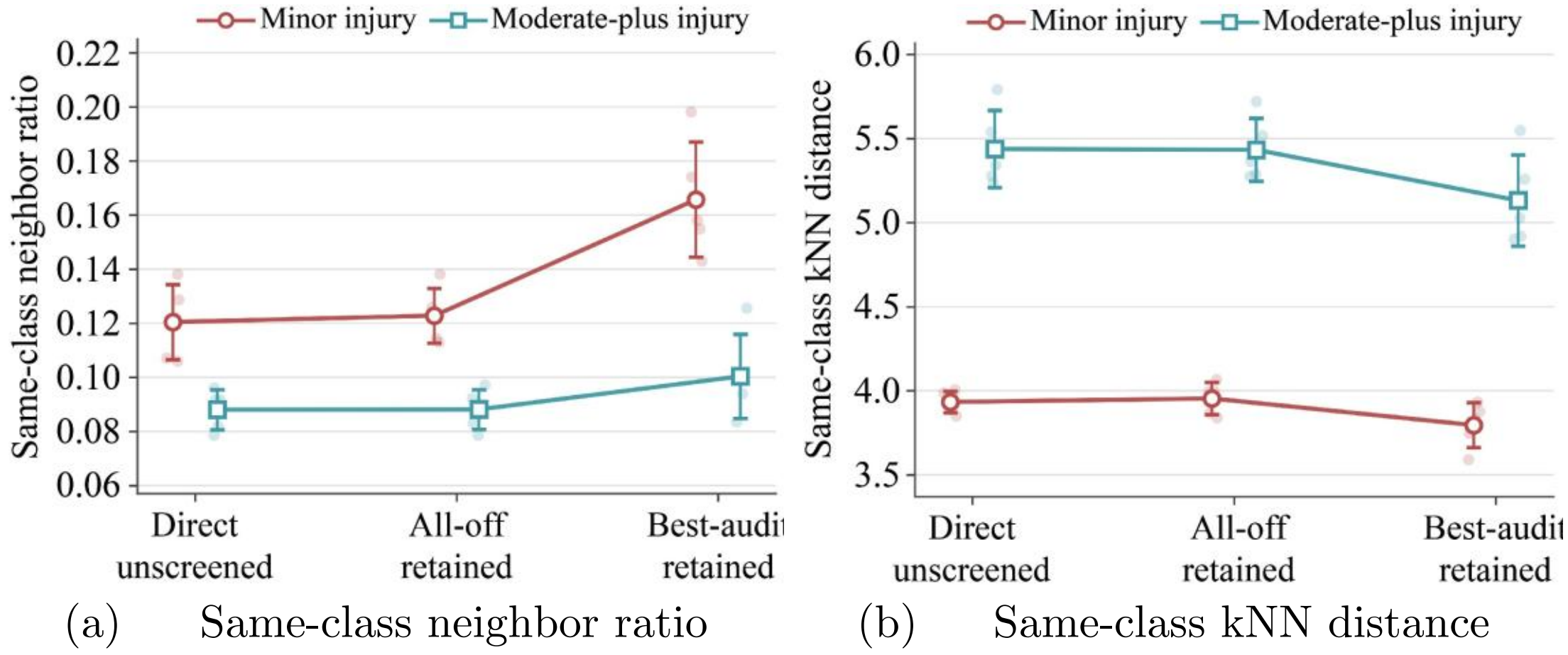

(a) Same-class neighbor ratio (b) Same-class kNN distance

**Fig. 3**. Local neighborhood support of accepted synthetic samples

For *Minor injury*, ECAS distinguishes itself from both references. Direct unscreened augmentation and all-off retention remain nearly unchanged: the neighbor ratio moves from 0.1205 to 0.1229, and the distance increases from 3.9330 to 3.9544. ECAS raises the ratio to 0.1657 and lowers the distance to 3.7957, placing the retained samples closer to real minor-injury neighborhoods.

The higher severity class follows the same direction despite greater sparsity. The neighbor ratio increases from about 0.088 to 0.1004, and the distance falls from about 5.44 to 5.13. This movement matters because *Moderate-plus injury* provides few real cases for learning; a retained synthetic sample must be locally defensible before it can usefully shape the class boundary. The all-off rule behaves much like unscreened augmentation, while ECAS changes which generated records are allowed into training.

### 5.3 Model interpretation

Although predictive performance provides a necessary benchmark, its value in ADS injury severity prediction remains limited if the model behavior cannot be interpreted in safety-relevant terms. For this reason, the analysis in this section examines the final model from two complementary perspectives. SHAP is used to identify the variables and feature values that contribute to class-specific predictions, while PDP analysis is used to assess how the average predicted probability changes when key features are varied over their observed categories.

#### *5.3.1 Global importance*

A clear split appears between the lower injury-severity classes and the highest severity class in the class-level SHAP rankings (Fig. 4). For *No injury* and *Minor injury*, the leading variables are concentrated in the immediate crash interaction. Crash-partner movement is the strongest contributor in both panels, while crash counterpart and subject-vehicle movement also remain in the leading group. Month is still visible, but the main information used by the model comes from the crash configuration itself: how the crash partner moved, what type of counterpart was involved, and what the ADS vehicle was doing before impact. Roadway and weather-related variables play a more limited role in these two classes.

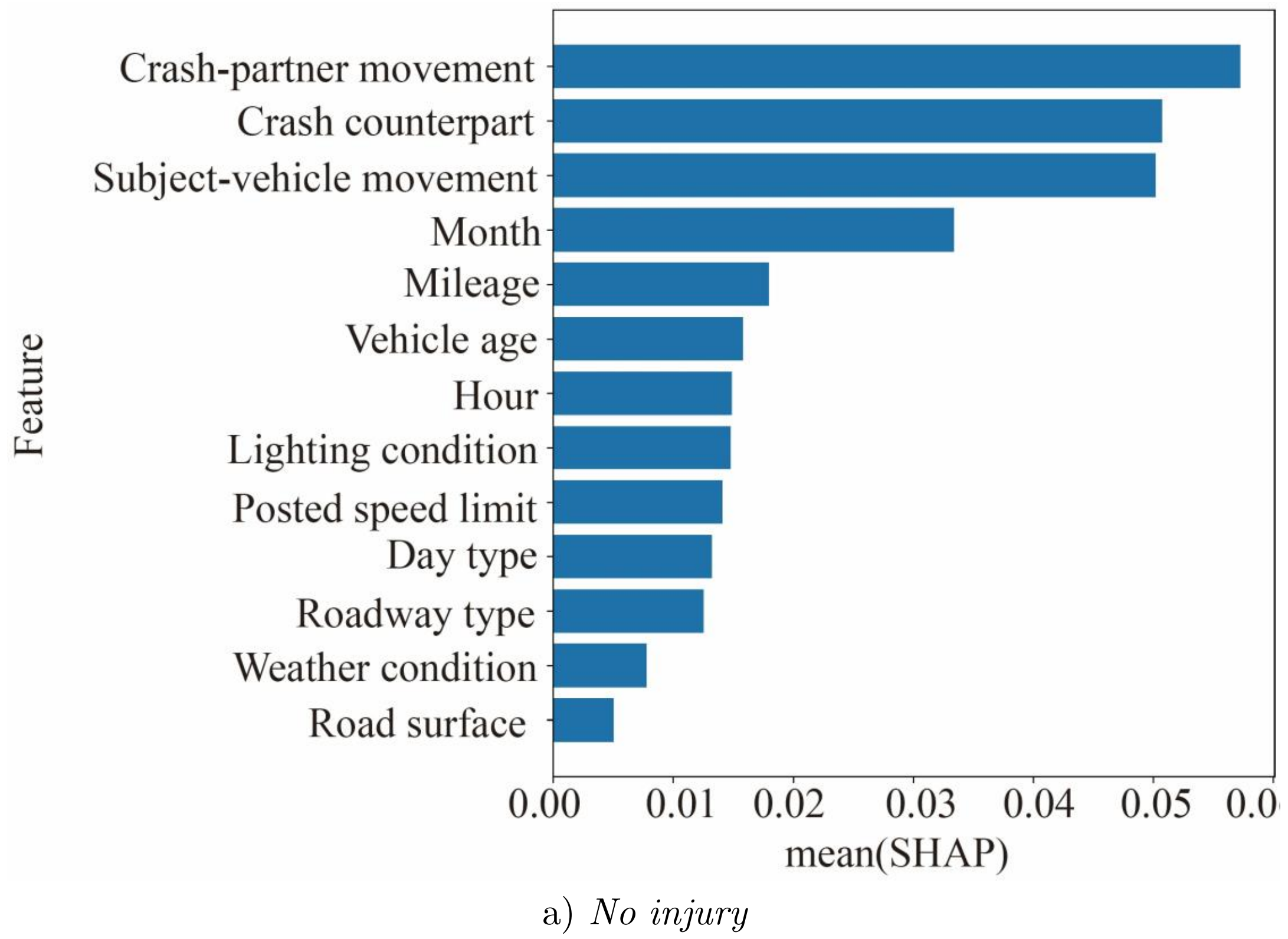

a) *No injury*

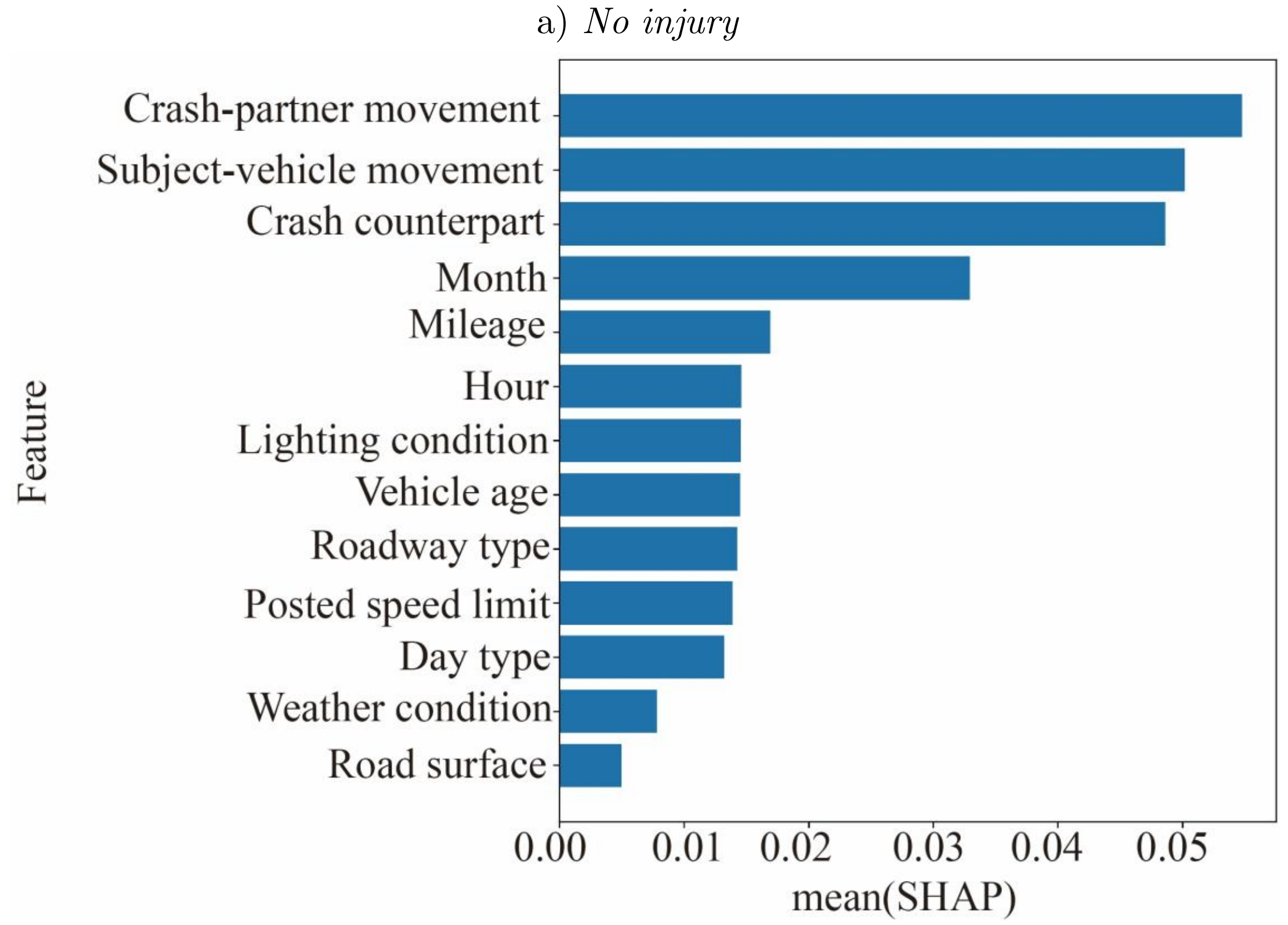

b) *Minor injury*

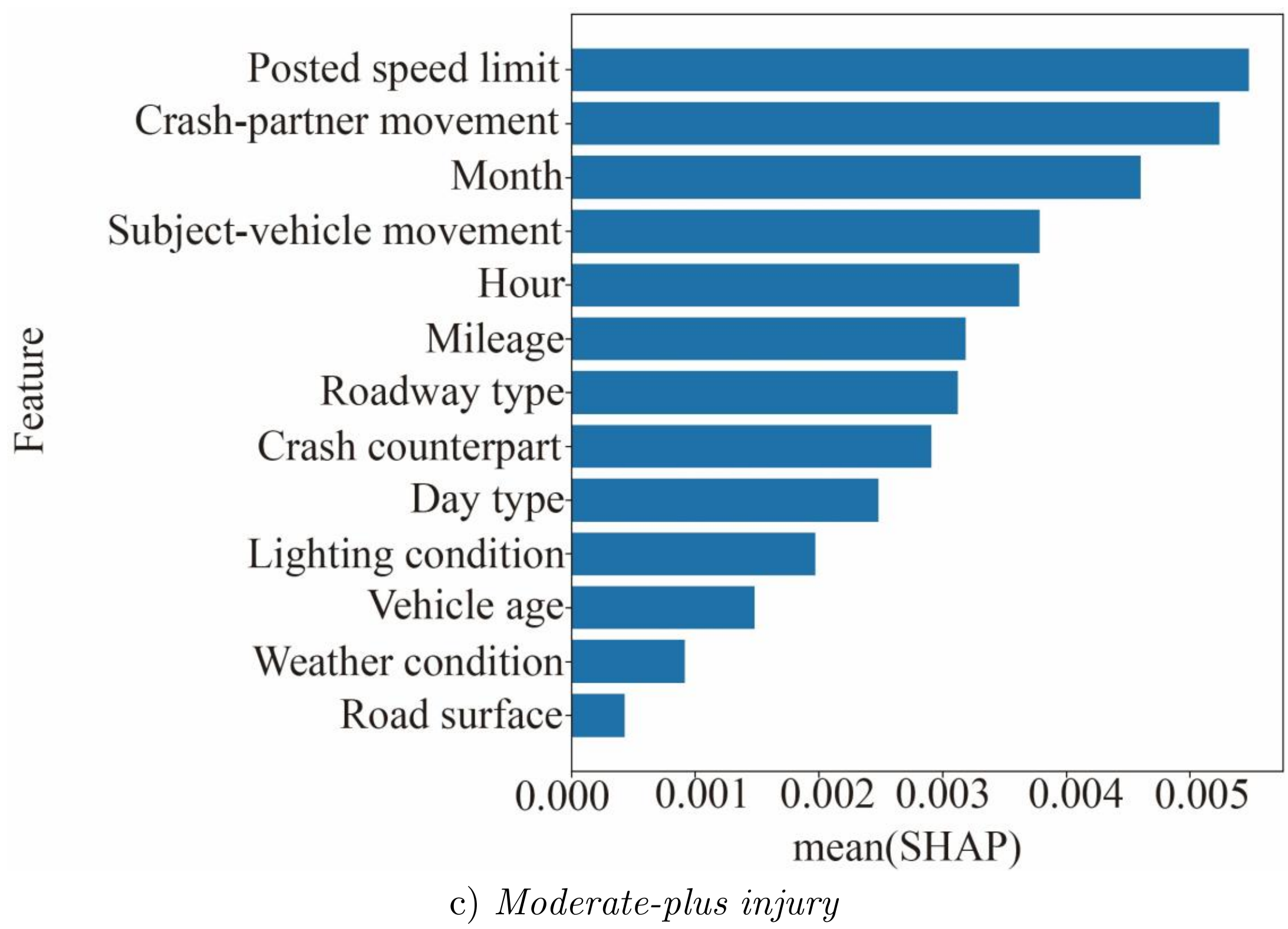

c) *Moderate-plus injury*

**Fig. 4.** Global SHAP importance across the three injury severity classes.

The ranking changes for *Moderate-plus injury*. Posted speed limit becomes the strongest contributor, and variables related to temporal context, roadway setting, and exposure occupy more prominent positions than in the two lower severity panels. Crash counterpart, which plays a central role for *No injury* and *Minor injury*, is less dominant here. The severe class is therefore not characterized only by the immediate conflict type. Its prediction draws more heavily on the operating context surrounding the crash, especially the speed environment and the conditions under which the ADS vehicle was operating. This pattern is also consistent with the scarcity of *Moderate-plus injury* cases, where common interaction features alone may not provide enough separation.

For ADS safety assessment, the difference between the three panels points to a more scenario-oriented interpretation of injury severity. Evaluation aiming at routine crash avoidance can reasonably emphasize conflict configuration and counterpart recognition, because these variables dominate the lower severity classes. Evaluation aiming at reducing more severe injuries needs a wider coverage of high-speed environments, roadway contexts with more demanding operating conditions, and time periods that may affect visibility, traffic composition, or available response margins.

*5.3.2 Main effects*

Global importance indicates which variables the model relies on most, but it does not show how specific values within those variables change class specific attribution. That distinction matters in ADS crash analysis, because the same variable can correspond to very different conflict settings once its internal categories are separated. The main-effect results therefore add a more operational layer of interpretation.

(1) *No injury*

Figure 5 first clarifies that *No injury* should not be treated simply as a residual majority class. Although this outcome dominates the sample, its SHAP profile still shows a coherent structure. The highest-ranked variables are crash-partner movement, crash counterpart, subject-vehicle movement, month, mileage, and vehicle age. This ranking indicates that the model assigns *No injury* mainly through the degree to which the crash interaction remained contained.

The strongest evidence comes from crash-partner movement. Backing/parking and departure/opposing have positive SHAP values for *No injury*, whereas straight movement has a negative value for this class. This contrast suggests that crashes involving more localized or constrained partner movement were more compatible with a non-injury outcome. In the fitted model, straight movement is treated as a more active conflict state than parking or backing categories, which is consistent with its lower *No injury* attribution. The same logic appears in subject-vehicle movement. Stopped/parked and several constrained movement categories remain on the positive side, whereas turning and straight movement have negative attributions. *No injury* therefore appears most likely when the interaction has limited kinetic and spatial development, rather than when either party continues to negotiate an active path through the crash scene.

Crash counterpart adds the clearest contrast in the *No injury* profile. Vulnerable user involvement has a strongly negative SHAP value for *No injury*, while other, heavy vehicle, and fixed object have positive values. This pattern indicates that the fitted model assigns lower *No injury* attribution to crashes involving exposed road users, rather than treating all low consequence crashes as the same prediction profile. A contact involving an exposed road user sharply reduces the *No injury* prediction, even if other conditions appear relatively ordinary. The positive values for fixed object and larger vehicle counterparts should be read as model associations, possibly reflecting contained contacts, parking related incidents, or limited damage encounters in the reported data.

The remaining variables refine this interpretation rather than replacing it. Month contributes to the *No injury* profile, with Nov and Dec showing positive tendencies and Feb and Mar shifting the prediction downward. Mileage also separates the class modestly: 1,000–10,000 miles and 10,000–50,000 miles are closer to the positive side, whereas >50,000 miles is negative. Vehicle age shows a similar secondary pattern, with 3–5 years and ≥6 years contributing more positively than 0–2 years. These effects function as contextual signals linked to exposure composition, operating maturity, or reporting conditions. They do not define *No injury* as strongly as the interaction variables do.

This class therefore marks the lower boundary of ADS crash severity rather than a residual category formed after injury cases are removed. It describes crashes in which contact occurs, but the movement configuration and counterpart type allow the event to remain within a low-consequence envelope.

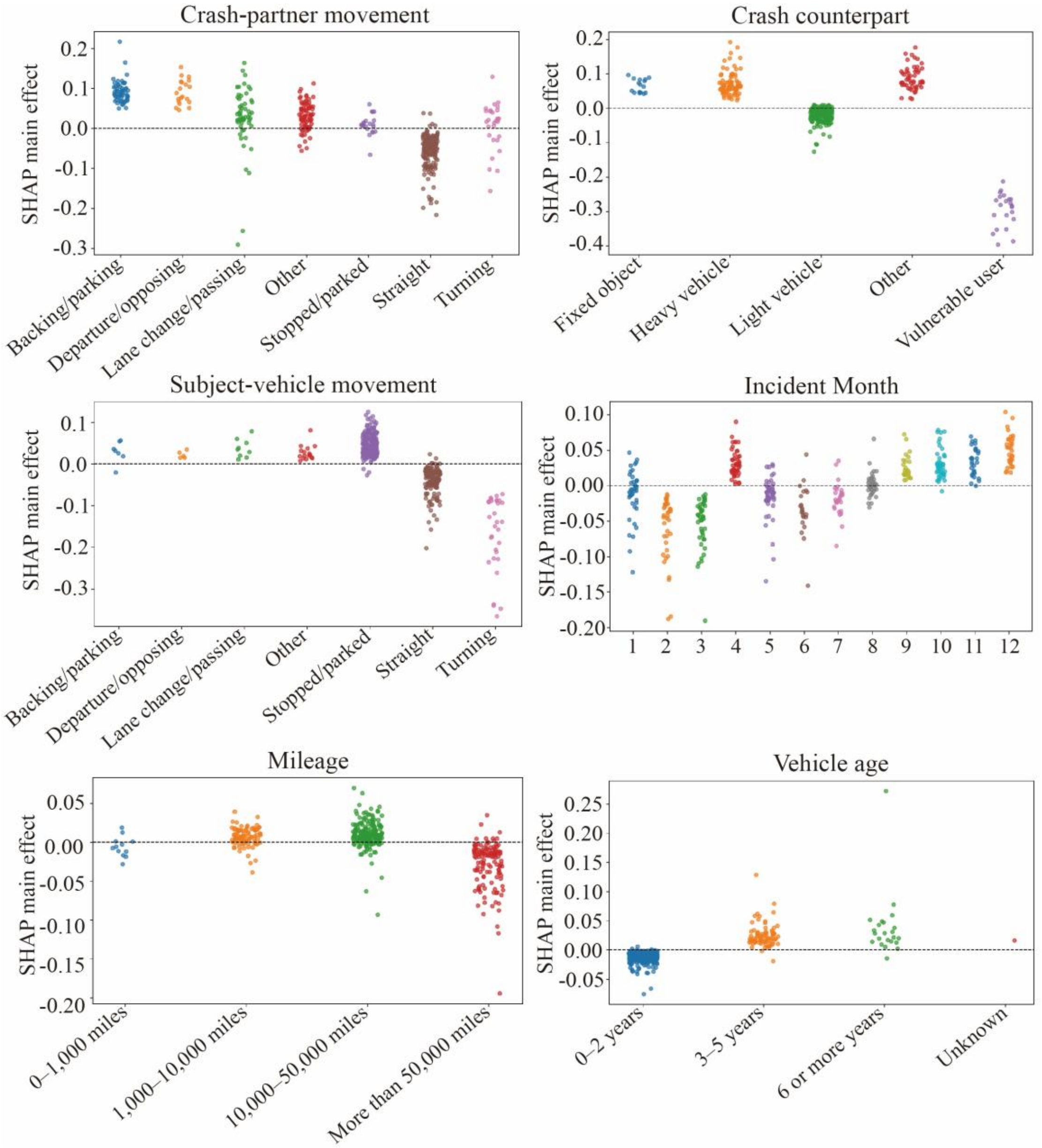

**Fig. 5.** Main-effect SHAP distributions for *No injury*.

(2) *Minor injury*

*Minor injury* has a different structure from the *No injury* panel, as shown in Fig. 6. Its three leading variables are crash-partner movement, subject-vehicle movement, and crash counterpart, placing the class in the immediate crash interaction rather than in general roadway context. This ordering weakens a simple ordinal reading of the outcome scale. *Minor injury* is not simply a midpoint between *No injury* and *Moderate-plus injury*; it is associated with active conflict configurations that differ from both contained *No injury* cases and the speed related profile of *Moderate-plus injury*.

The two movement variables provide the first piece of evidence for this interpretation. In the crash-partner movement profile, straight travel contributes positively to *Minor injury*, whereas backing/parking and departure/opposing contribute negatively. In the subject-vehicle movement profile, turning shows the strongest positive contribution, and straight travel is also on the positive side. By contrast, stopped/parked and lane change/passing reduce the tendency toward this class. These results suggest that *Minor injury* is most closely associated with conflicts in which one or both parties are still actively moving through the conflict space. More constrained movement categories, such as backing/parking, departure/opposing, and stopped/parked, have lower or negative attributions for Minor injury, so the fitted model treats them as less characteristic of this outcome than straight or turning movements.

Crash counterpart gives this pattern a clearer safety interpretation. Vulnerable-user involvement has the strongest positive SHAP contribution among the counterpart categories, while other, heavy vehicle, and fixed object are negative. This contrast suggests that the *Minor injury* boundary is especially sensitive to conflicts involving pedestrians, cyclists, or other exposed road users. In such encounters, the same vehicle movement may carry greater injury potential because the counterpart has limited physical protection and may follow less predictable trajectories. This pattern is consistent with the broader ADS safety literature, where vulnerable-road-user interactions are often analyzed as a critical scenario family rather than as a simple counterpart category.

The remaining variables function more as modifiers of this interaction pattern than as the main explanation of *Minor injury*. Month appears in the top six ranking, with Feb, Mar, and Jun contributing positively and several later months contributing negatively. Mileage also adds a secondary distinction, with >50,000 miles showing a positive attribution for *Minor injury*. Hour is weaker, and its category-level differences are small. These patterns may reflect exposure composition, operating area, or reporting conditions, but their attribution is weaker than that of movement and counterpart variables. *Minor injury* in ADS crashes is best understood as an active-conflict outcome, especially those involving vulnerable users, rather than by roadway type, lighting condition, or posted speed limit.

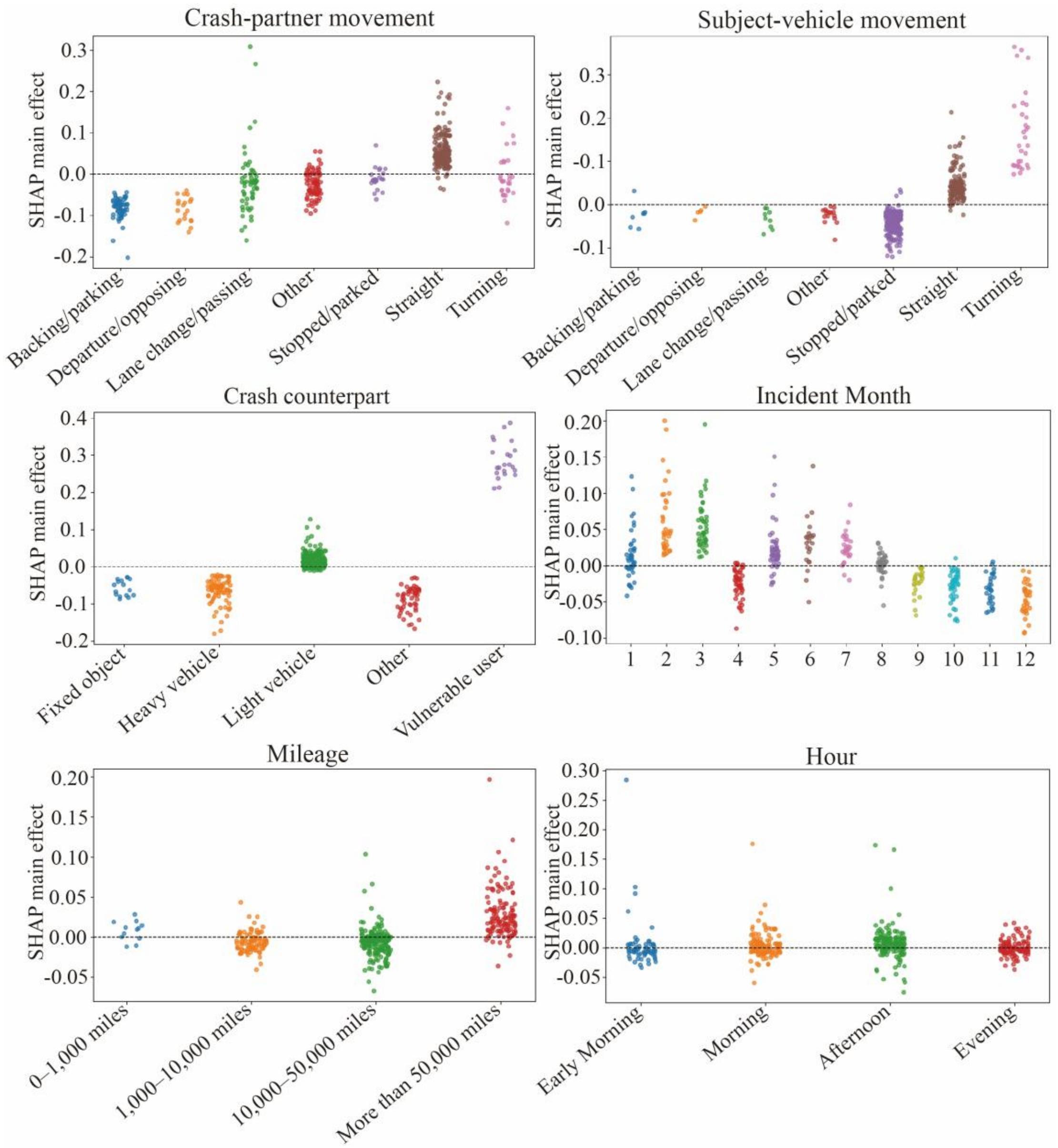

**Fig. 6**. Main-effect SHAP distributions for *Minor injury.*

(3) *Moderate-plus injury*

The explanatory profile of *Moderate-plus injury* differs substantially from the two lower severity classes. Whereas *No injury* and *Minor injury* are mainly separated by crash interaction variables, the rarest class shows a more diffuse and context-dependent structure, as shown in Fig. 7. Posted speed limit ranks first, followed by crash-partner movement, month, subject-vehicle movement, hour, and mileage. The absolute SHAP magnitudes are much smaller than those observed for the other two classes, indicating that the model captures a less sharply separated severe-class pattern. This result is consistent with the empirical difficulty of learning from a class that accounts for only a very small share of the ADS crash records.

The most interpretable signal comes from posted speed limit. The categories 40–45 mph and ≥50 mph have positive SHAP values for *Moderate-plus injury*, whereas 0–20 mph and 25–35 mph have negative values. This value-level contrast places the severe-class attribution closer to the speed environment than to the fine-grained configuration of the crash interaction. Unlike *Minor injury*, where exposed road users and active movement states form a clearer attribution pattern, *Moderate-plus injury* is more closely aligned with a higher-speed operating context. This does not imply that speed alone determines injury severity; it indicates that the prediction function assigns higher severe-class attribution to higher speed regimes.

The movement variables play a different role here than they do for the two lower classes. In the No injury profile, constrained movements have positive attribution; in the Minor injury profile, active movements and vulnerable-user encounters become more prominent. For *Moderate-plus injury*, however, crash-partner movement and subject-vehicle movement do not produce a comparably sharp separation. Stopped/parked as a crash-partner movement remains on the positive side, and turning as a subject-vehicle movement is close to the positive side, but the remaining categories cluster near zero or on the negative side. This compressed movement profile indicates that the highest injury category is not represented as a simple extension of the lower-class interaction pattern. The model appears to rely less on whether the crash involved a particular maneuver and more on whether the crash occurred under conditions where consequences could escalate.

The secondary variables mainly locate *Moderate-plus injury* within the operating context. Month enters the top ranking, with Jun showing the clearest positive tendency, but the pattern is too limited to support a seasonal explanation. Hour contributes little beyond this context, although afternoon is higher than the other periods. Mileage is highest for >50,000 miles, which may reflect accumulated exposure or deployment context rather than a direct injury mechanism. These variables help position the crash within a broader operating context, but they do not define the class in the way that crash counterpart and movement variables define *Minor injury*.

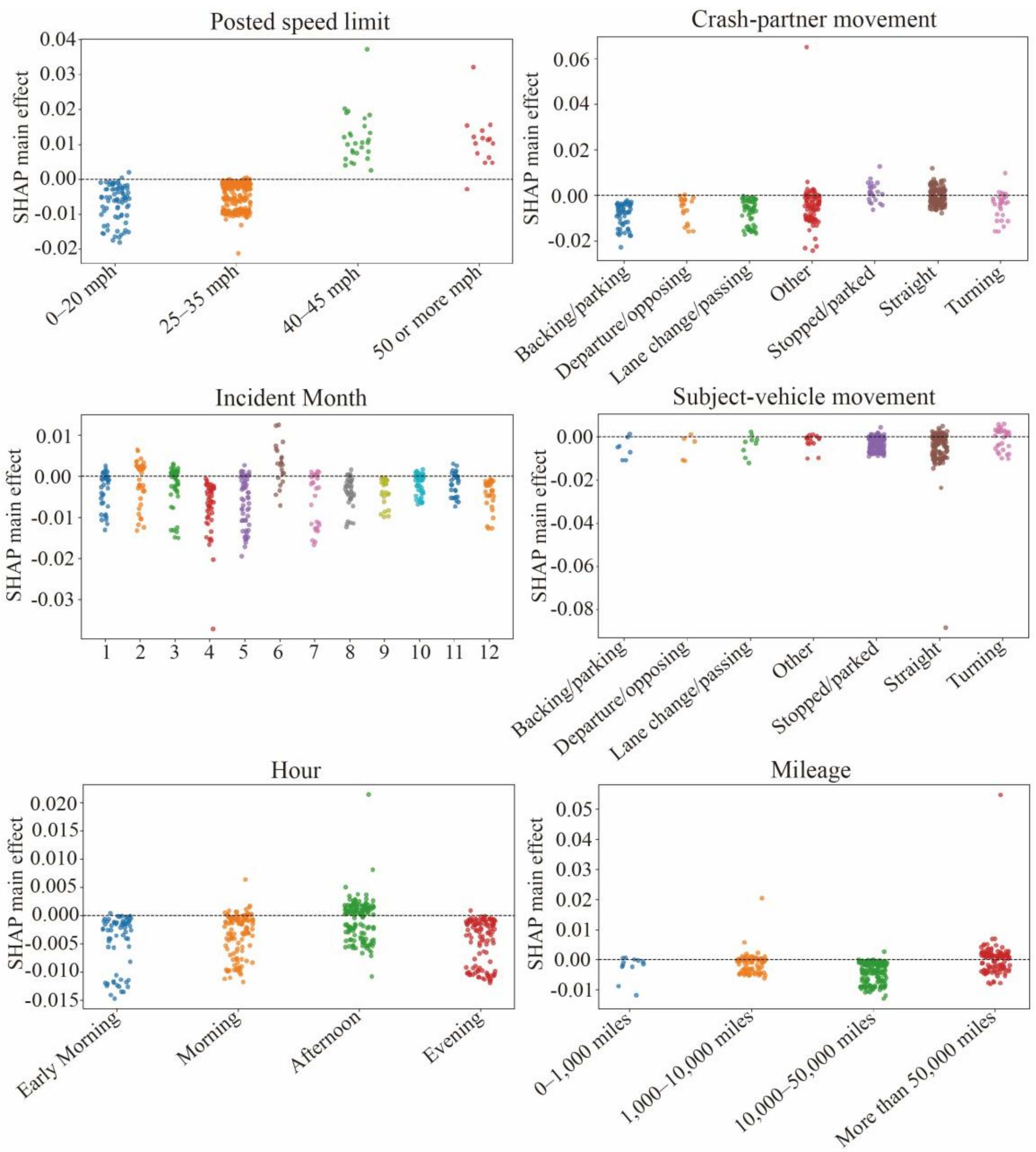

**Fig. 7.** Main-effect SHAP distributions for *Moderate-plus injury.*

### *5.3.3 PDPs*

Partial dependence adds a probability-scale check to the SHAP results. In Figs. 8, 9, and 10, each PDP fixes one SHAP-selected variable at a given value and records the mean probability assigned to the target injury class by the fitted model. The gray reference line is the original test-set mean probability for that class before perturbation. Since these variables are categorical or discretized, the connected line is used to guide comparison across value levels, not to imply a continuous change between categories.

For *No injury*, the most informative pattern is not the high baseline itself, but the few conditions that pull the probability away from it. This is most visible in Fig. 8. Crash counterpart gives the sharpest separation: heavy vehicle, fixed object, and other remain above 0.915, whereas a vulnerable user lowers the *No injury* probability to 0.588. Subject-vehicle movement shows a similar loss of containment. Turning reduces the probability to 0.701, and straight travel also falls below the baseline. By contrast, backing/parking, departure/opposing, lane change/passing, and stopped/parked stay close to the upper range. The model therefore assigns lower *No injury* probability to exposed road user and active maneuvering categories, while bounded movement categories remain closer to the upper probability range.

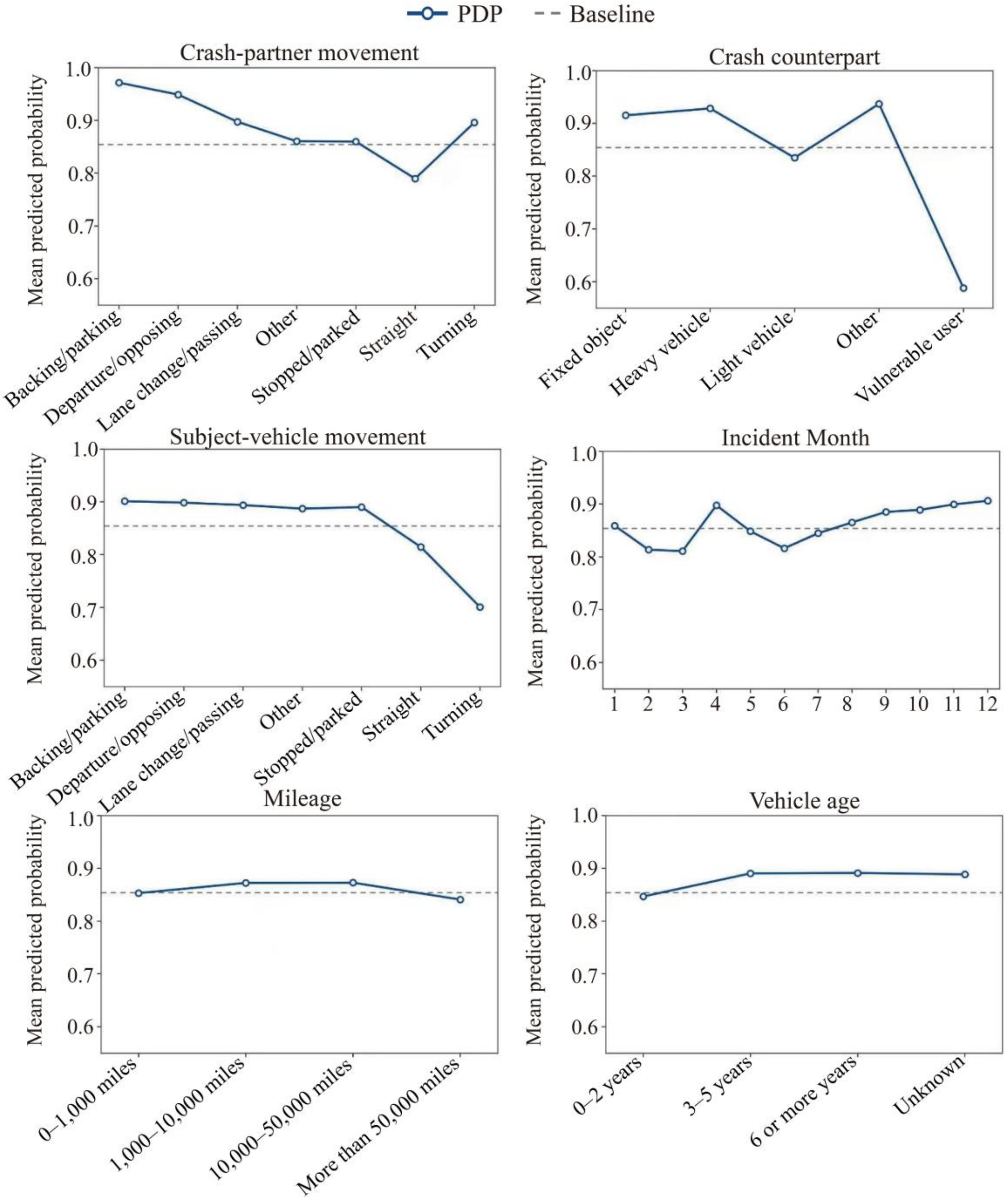

**Fig. 8.** PDPs for *No injury*.

*Minor injury* has a distinct probability signature rather than a simple intermediate position on the severity scale. The profiles in Fig. 9 rise most strongly when the conflict involves an exposed counterpart or active vehicle movement. Vulnerable user increases the *Minor injury* probability to 0.387, far above the class baseline of 0.135. Turning by the subject vehicle reaches 0.286, and straight movement by either party also lies above the baseline. These responses match the SHAP evidence but give a clearer probability interpretation: *Minor injury* is most strongly associated with conflicts that are still dynamically active at the moment of impact. Month and mileage provide secondary variation, with higher values in Feb, Mar, and Jun and with $>$50,000 miles, whereas the changing hour across the day has little effect. The weak hour profile helps keep the interpretation centered on interaction structure rather than broad temporal exposure.

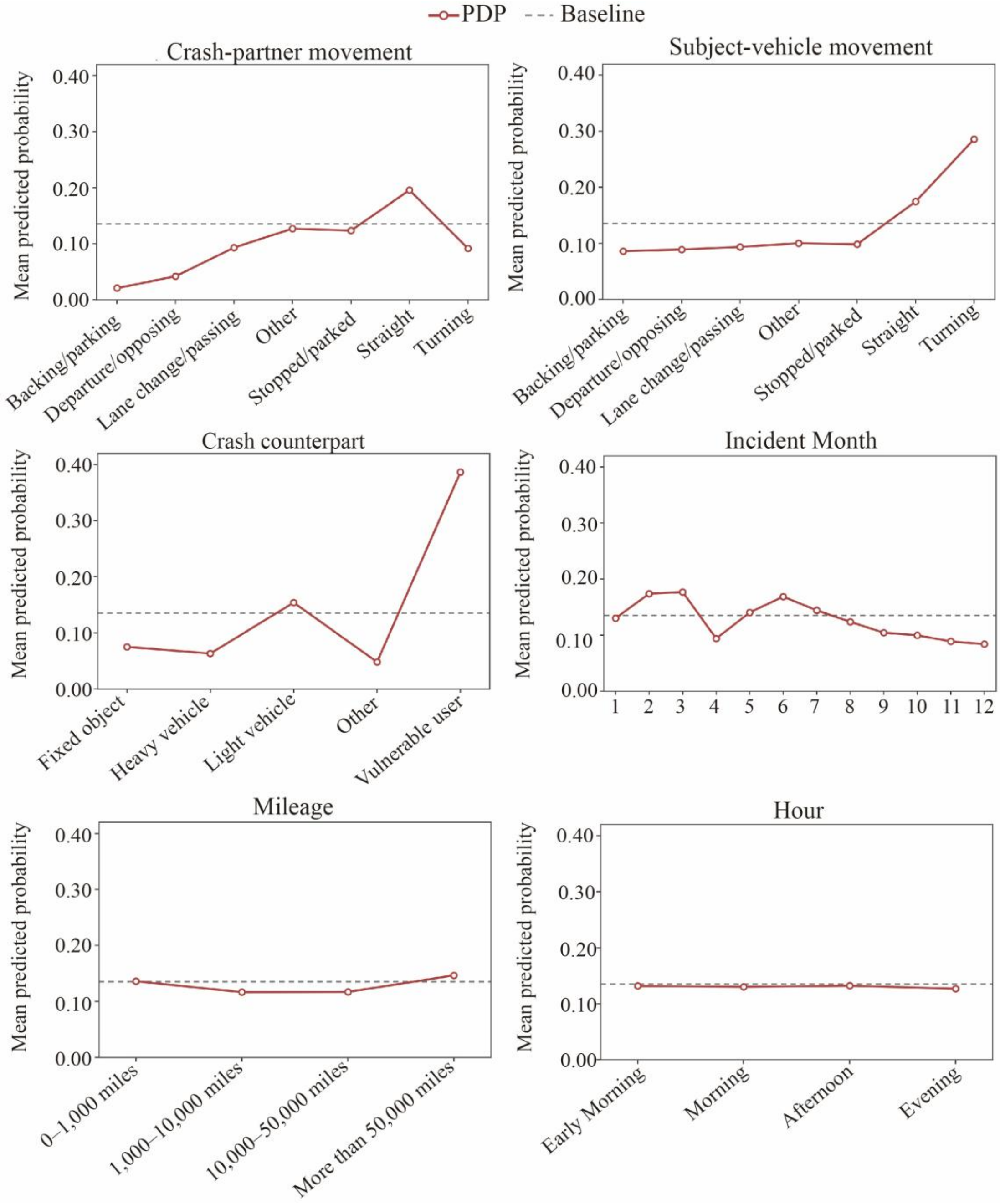

**Fig. 9.** PDPs for *Minor injury.*

The *Moderate-plus injury* result is narrower. In Fig. 10, posted speed limit shows the largest change on the severe class probability scale. The predicted probability stays near 0.010 under 0–20 mph and 25–35 mph, then rises to 0.024 at 40–45 mph and 0.027 at ≥50 mph. This speed response is consistent with crash energy reasoning, although the absolute probability remains low. The other profiles fluctuate close to the baseline: crash-partner movement, subject-vehicle movement, hour, month, and mileage only make small differences to the severe-class probability. Because the hard-label performance for *Moderate-plus injury* is zero, these curves provide probability scale evidence rather than evidence of stable severe-injury identification. The model assigns higher severe-class propensity to higher speed regimes, but it still does not convert that propensity into reliable hard-label recognition.

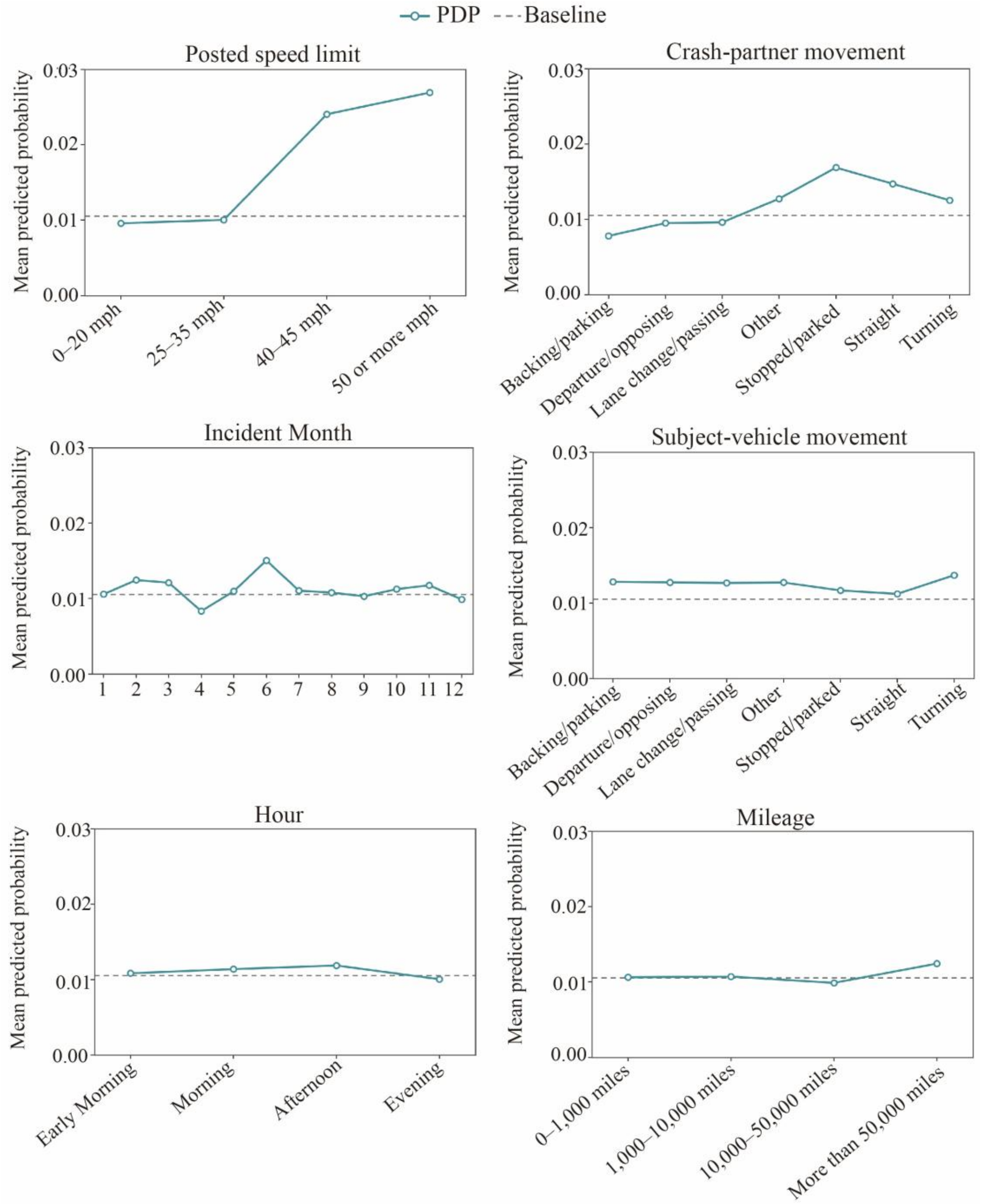

**Fig. 10**. PDPs for *Moderate-plus injury*.

The two interpretive views separate attribution from probability response. SHAP identifies the variables that structure each class-specific prediction; PDP analysis shows how the model's assigned probability changes when those variables take specific values. Both views place crash counterpart and movement configuration at the center of the model boundary between *No injury* and *Minor* *injury*. The severe-class boundary is different. It is less supported by a broad interaction pattern and more by a localized response to posted speed limit, with secondary changes from movement, time, and mileage. This distinction supports a more defensible inference: the framework reveals meaningful severity-related probability responses, while the rare *Moderate-plus injury* class remains constrained by data sparsity and weak decision-boundary formation.

# 6. Discussion

## 6.1 Expert committee audit screening

The methodological value of ECAS lies in changing the unit of decision in ADS crash augmentation. Conventional imbalance treatments usually decide how many minority observations to add, reweight, or resample. ECAS instead asks which generated records are credible enough to enter training. This distinction matters for ADS crash data because higher injury-severity classes are sparse, feature combinations are heterogeneous, and a small number of generated records can influence the learned boundary. Cost-sensitive learning, SMOTE, random oversampling, class weighting, and ensemble classifiers help reduce majority-class dominance (Zhu and Meng, 2022; Channamallu et al., 2024, 2025), but they do not by themselves determine whether a generated crash is well supported by observed crashes.

The expert committee provides the audit screening layer. Label support checks whether the intended class is plausible under models trained only on real data. Boundary separation checks whether the candidate is distinguishable from competing classes. Committee agreement tests whether different model families give compatible judgments. Local plausibility brings the candidate back to the empirical crash distribution by examining neighborhood support. These four channels do not act as independent indicators. They form a record acceptance system in which a generated sample must be defensible as a class member, as a decision point, and as a local observation.

The class-wise Pareto step keeps the screening mechanism from becoming a manually weighted score. Candidate samples are compared within their generated class, which reduces scale effects across severity levels. Pareto non-dominated sorting then gives priority to samples that are not clearly inferior across the enabled evidence channels. This design is suited to safety-related prediction, where assigning a fixed empirical weight to confidence, margin, consistency, and density would be difficult to defend. The selected configuration, *audit_c1m1s0d1*, also shows why ECAS should not be read as a stricter filter that always benefits from more channels. The full evidence setting did not perform best. The effective decision was selective acceptance, not maximum filtering.

The best-performing configuration combines confidence, margin, and density while disabling consistency. This pattern is methodologically meaningful. Confidence ensures that the generated label is supported by the expert committee in probabilistic terms. Margin further requires the candidate sample to be separated from the strongest competing class, which helps avoid synthetic samples located near unstable decision boundaries. Density adds a data-structure constraint by favoring samples located close to locally supported regions of the real crash distribution. Together, these three channels screen generated samples from three complementary perspectives: class support, boundary clarity, and local plausibility.

The exclusion of consistency from the best configuration does not mean that committee agreement is conceptually irrelevant. Rather, it suggests that hard-label agreement may be too restrictive under severe class imbalance and small minority-class samples. Unlike confidence and margin, which preserve probabilistic information, consistency reduces expert outputs to discrete decisions. In sparse minority regions, different experts may reasonably disagree for plausible samples because they rely on different inductive biases and because the minority-class boundary is weakly supported by real observations. Penalizing such disagreement may remove informative minority samples, especially those near meaningful but underrepresented decision regions. This explains why the full configuration performs slightly worse than the confidence-margin-density configuration. For ADS crash injury severity prediction, credible augmentation therefore requires a balance between screening out weak synthetic samples and retaining sufficient minority-class variation.

The neighborhood results ground this interpretation at the data level. ECAS-accepted samples occupy real minority-crash neighborhoods more closely than samples retained without audit evidence. Therefore, the observed gain is tied to which generated records enter training, not only to how many minority records are added. Future ADS injury prediction studies should evaluate augmentation by sample support as well as sample quantity. ECAS makes this acceptance decision explicit without collapsing the evidence channels into a manually weighted score.

### 6.2 Injury severity patterns of ADS crashes

The present findings indicate that ADS crash injury severity predictions are most strongly associated with the joint configuration of movement, counterpart type, and operating context, rather than by any single factor alone. Earlier studies using California DMV crash reports have linked crash severity to vehicle movement, collision type, facility type, lighting, and driving mode (Wang and Li, 2019). Studies using crash narratives have reached a similar conclusion from a different data type, showing that unstructured descriptions contain information on pre-crash behavior, roadway context, and crash participants that is not fully captured by conventional summary variables (Lee et al., 2023). The contribution of the present study lies in separating these relationships by injury severity class. *No injury* and *Minor injury* were mainly distinguished by interaction-related variables, especially crash-partner movement, subject-vehicle movement, and crash counterpart. *Moderate-plus injury*, in contrast, was more sensitive to posted speed limit and roadway type. The model did not treat injury severity as a single continuum driven by uniformly stronger versions of the same factors, but separated the classes through different severity boundaries.

A clearer picture emerges when the lower severity classes are examined together. *No injury* was most compatible with constrained or localized encounters, while *Minor injury* appeared when the crash setting involved more active movement and a less stable interaction boundary. This finding extends prior work in two ways. Wang and Li (2019) and Zhu and Meng (2022) identified movement before collision as an important contributor to AV crash severity, but their analyses focused more on overall severity classification or binary injury classification. By separating *No injury* from *Minor injury*, the analysis indicates that movement variables contribute differently to the fitted prediction function across outcome boundaries. Stopped or parked states supported the *No injury* prediction, whereas straight and turning movements became more relevant once the model moved toward *Minor injury*. The operational implication is more specific. For ADS testing, the critical question is not only whether the vehicle detects a hazard, but whether it can interpret the evolving motion of other road users early enough to keep the event within a no-injury boundary.

The role of vulnerable road users is one of the most policy-relevant findings. Kutela et al. (2022) showed that vulnerable-road-user involvement in AV crashes was closely related to crosswalks, intersections, traffic signals, and AV movements. Li et al. (2024) later used topic modeling and explainable XGBoost to show that crash narrative topics involving vulnerable road users had substantial influence on severity prediction. The present analysis reaches a comparable conclusion through structured ADS variables rather than text-derived topics. Vulnerable user was associated with lower predicted probability of *No injury* and higher predicted probability of *Minor injury*, suggesting that the fitted model treated exposed road user involvement as an important discriminator between these two outcome classes. This result is not surprising from a mechanism perspective. Pedestrians, cyclists, and other exposed users have limited physical protection, their trajectories may be harder for other road users to infer, and many of their conflicts with ADS vehicles unfold in shared urban space. In such scenarios, conservative speed choice, intent prediction, and external communication become part of the same safety problem rather than separate design issues.

The pattern for *Moderate-plus injury* pointed to a different part of the ADS safety envelope. Here, movement variables were less dominant, while posted speed limit and roadway type became more informative. This finding is consistent with studies that have emphasized highway- and speed-related conditions in AV crash severity. Kuo et al. (2024) reported that speed limit and roadway features were associated with AV crash severity after addressing small-sample and imbalance issues. Ding et al. (2024) also found that roadway type and exposure conditions played a role in injury severity levels involving vehicles with driving automation. In the present model, 40–45 mph and ≥50 mph were associated with higher *Moderate-plus* attribution, and highway/freeway made a positive contribution. This result indicates a stronger probability response under higher-speed and less forgiving roadway settings, supporting separate evaluation of low-speed urban operations and higher-speed operating domains.

Lighting and temporal variables function as contextual indicators rather than direct causal mechanisms. Daylight was more compatible with *Minor injury*, and afternoon operation showed a stronger tendency toward *Moderate-plus injury* than other time periods. Similar time-dependent and context-dependent effects have been reported in recent scenario-based SHAP studies of Level 4 AV crashes (Wang et al., 2026). These variables may represent exposure conditions that are not fully captured by the structured features, such as traffic density, road user mix, and the frequency of interaction with pedestrians, cyclists, or turning vehicles. Their role is therefore best understood as situating the crash within a broader operating context. This interpretation also explains why time and lighting should be preserved in ADS crash reporting, even when they are not sufficient explanatory variables on their own.

ADS injury severity is therefore better described through safety boundaries than through isolated risk factors. The *No injury* boundary remains strongest when conflicts are localized and controllable. The *Minor injury* boundary appears when the ADS must manage moving interactions, especially with vulnerable road users in ordinary urban traffic. The upper severity boundary is tied more closely to speed environment and roadway context, where reduced correction time and lower tolerance for error can amplify crash consequences.

**6.3 Implications on safety management**

The findings suggest that ADS safety oversight should move from aggregate event counting toward scenario-based monitoring. A crash record is most useful when it preserves the operating context in which the event occurred, including pre-crash movement, counterpart behavior, roadway type, posted speed limit, lighting, time of day, and operating domain. These fields allow agencies and operators to distinguish routine low-consequence contacts from recurring scenarios that may indicate a weakening safety boundary. For example, repeated low-speed contacts in constrained maneuvering settings raise a different concern from repeated interactions with moving road users in ordinary urban traffic, and crashes in higher-speed roadway contexts may deserve dedicated review even when their frequency is low.

For deployment management, the practical focus should be on where the system begins to operate with reduced tolerance for error. This requires monitoring not only final injury labels, but also the conditions under which predicted outcomes shift across injury boundaries. Predictive models can support this process, but their use in safety governance should remain transparent. When augmented samples are used, agencies and developers should be able to examine how minority severity information entered the learning process, which samples were retained, and which operating conditions contributed to the prediction. In this respect, model performance should be reviewed together with ECAS acceptance records, sensitivity checks, class-specific explanations, and probability response profiles, rather than being judged only by a single accuracy metric.

The same logic applies to scenario design. Testing programs should not simply expand the number of cases in a generic scenario catalog. They should identify scenario gaps that correspond to fragile outcome boundaries and then vary the elements that affect task difficulty, such as counterpart type, path timing, speed choice, visibility, roadway setting, and yielding behavior. These implications should be understood as guidance for safety learning rather than final regulatory rules. Deployment decisions still require exposure information, operational design-domain documentation, simulation evidence, and post-deployment monitoring. The value of the present framework is that it links prediction, explanation, and audit evidence, thereby helping agencies and manufacturers decide which crash records require closer review, which scenario families should be expanded in testing, and which reporting fields should be improved for future ADS safety assessment.

### 6.4 Limitations and future studies

Three limitations define the scope of the findings. The analysis relies on reported ADS crash records, so the results depend on the completeness and consistency of the original reports. Some variables remain coarse, missing, or grouped into broad categories. This restricts the ability to separate finer operating conditions, including traffic density, exact conflict geometry, road user visibility, and the ADS state before impact.

The rarity of higher injury-severity classes remains a second constraint. ECAS improves sample acceptance under sparse data, but sparse observations still limit the stability of class-specific patterns, especially for *Moderate-plus injury*. The results describe model sensitivity under the available records, not a complete account of ADS injury mechanisms. Larger multiyear datasets, exposure measures, and external validation across deployment areas and operating entities are needed to test the stability of these patterns.

A third limitation concerns model scope. The present study evaluates ECAS under the selected NF-TabPFN backbone and four audit evidence channels. The relative roles of label support, boundary separation, committee agreement, and local plausibility may change with feature representation, dataset size, model family, and imbalance level. Future work can test the framework with alternative learners, additional evidence channels, and external ADS crash datasets.

## 7. Conclusions

This study developed ECAS for ADS-crash injury-severity prediction using reported crash records from the NHTSA Standing General Order data. The analysis focused on *No injury*, *Minor injury*, and *Moderate-plus injury* classes, and addressed two connected problems: learning from sparse imbalanced injury-severity classes and interpreting the learned decision structure for safety assessment. ECAS places an audit screening layer between synthetic sample generation and model training. Candidate samples are evaluated through label support, boundary separation, committee agreement, and local plausibility before they enter training. We demonstrated that credibility-aware sample acceptance improves the reliability of injury severity prediction in ADS crash data, even when working with limited real-world observations. By prioritizing the inclusion of well-supported synthetic samples, the approach enables more effective learning from sparse data, potentially accelerating the assessment and monitoring of ADS safety without requiring large-scale data collection.

The results indicate that the quality of synthetic data played an important role in model performance. The configuration achieving the highest performance metrics incorporated only a subset of the available audit evidence channels, showing that additional channels do not automatically improve generalization. The neighborhood analysis provides the key data-level check: ECAS-accepted samples had stronger support from real minority crashes than direct unscreened and all-off retained samples.

The interpretation results also separate the three injury severity classes by scenario structure. *No injury* was linked mainly to contained encounters, *Minor injury* to active movement conflicts and vulnerable-road-user involvement, and *Moderate-plus injury* to speed and roadway context. These model-identified boundaries suggest that ADS safety testing should cover different scenario families rather than treating all injury levels as stronger or weaker expressions of one mechanism.

The study contributes an expert-committee sample acceptance framework for imbalanced ADS crash learning and an interpretable account of how crash conditions relate to injury severity. As ADS crash reporting becomes more complete, combining ECAS with class-specific interpretation can support more targeted scenario design, transparent safety monitoring, and evidence-based regulatory review.

**Data Availability Statement:** The raw crash data analyzed in this study are publicly available from the National Highway Traffic Safety Administration (NHTSA) Standing General Order crash reporting portal at https://www.nhtsa.gov/laws-regulations/standing-general-order-crash-reporting.